\documentclass[conference]{IEEEtran}
\IEEEoverridecommandlockouts
\usepackage{amsmath,amssymb,amsfonts}
\usepackage{algorithmic}
\usepackage{graphicx}
\usepackage{textcomp}
\usepackage{xcolor}
\usepackage[utf8]{inputenc} % allow utf-8 input
\usepackage[T1]{fontenc}    % use 8-bit T1 fonts
\usepackage{hyperref}       % hyperlinks
\usepackage{url}            % simple URL typesetting
\usepackage{booktabs}       % professional-quality tables
\usepackage{nicefrac}       % compact symbols for 1/2, etc.
\usepackage{microtype}      % microtypography
\usepackage{xcolor}         % colors
\usepackage{bbm}         % for math symbols
\usepackage[breakable]{tcolorbox}
\usepackage{orcidlink}
\usepackage[style=ieee, backend=biber]{biblatex}
\usepackage{multirow}       % for using minipages and resizing tables
\DeclareMathOperator*{\argmax}{arg\,max}

\newcommand{\tavfull}{\textit{Texture Association Value}}
\newcommand{\tav}{\textit{TAV}}
\newcommand{\codeurl}{\url{https://github.com/blainehoak/err-on-textures}}

\def\BibTeX{{\rm B\kern-.05em{\sc i\kern-.025em b}\kern-.08em
    T\kern-.1667em\lower.7ex\hbox{E}\kern-.125emX}}
\begin{document}

\title{Err on the Side of Texture:\\Texture Bias on Real Data}

\author{\IEEEauthorblockN{Blaine Hoak\orcidlink{0000-0003-2960-0686}}
\IEEEauthorblockA{
\textit{University of Wisconsin-Madison}\\
\textit{Department of Computer Sciences} \\
bhoak@cs.wisc.edu}
\and
\IEEEauthorblockN{Ryan Sheatsley\orcidlink{0000-0001-8447-602X}}
\IEEEauthorblockA{
\textit{University of Wisconsin-Madison}\\
\textit{Department of Computer Sciences} \\
sheatsley@wisc.edu}
\and
\IEEEauthorblockN{Patrick McDaniel\orcidlink{0000-0003-2091-7484}}
\IEEEauthorblockA{
\textit{University of Wisconsin-Madison}\\
\textit{Department of Computer Sciences} \\
mcdaniel@cs.wisc.edu}
}

% \author{\IEEEauthorblockN{1\textsuperscript{st} Given Name Surname}
% \IEEEauthorblockA{\textit{dept. name of organization (of Aff.)} \\
% \textit{name of organization (of Aff.)}\\
% City, Country \\
% email address or ORCID}
% \and
% \IEEEauthorblockN{2\textsuperscript{nd} Given Name Surname}
% \IEEEauthorblockA{\textit{dept. name of organization (of Aff.)} \\
% \textit{name of organization (of Aff.)}\\
% City, Country \\
% email address or ORCID}
% }

\maketitle

\begin{abstract}

Bias significantly undermines both the accuracy and trustworthiness of machine learning models. To date, one of the strongest biases observed in image classification models is texture bias---where models overly rely on texture information rather than shape information. Yet, existing approaches for measuring and mitigating texture bias have not been able to capture how textures impact model robustness in real-world settings. In this work, we introduce the \tavfull{} (\tav{}), a novel metric that quantifies how strongly models rely on the presence of specific textures when classifying objects. Leveraging \tav{}, we demonstrate that model accuracy and robustness are heavily influenced by texture. Our results show that texture bias explains the existence of natural adversarial examples, where over 90\% of these samples contain textures that are misaligned with the learned texture of their true label, resulting in confident mispredictions. 

\end{abstract}

% \begin{IEEEkeywords}
% texture bias, model bias, natural adversarial examples
% \end{IEEEkeywords}

\section{Introduction}

Bias serves as one of the core contributors of poor accuracy and lack of trustworthiness in machine learning models. One of the strongest biases observed in image classification models to date is texture bias~\cite{geirhos_imagenet-trained_2019, ballester_performance_2016, brendel_approximating_2019}---where models more strongly rely on the presence of textures, or repeated patterns, when classifying images. This intriguing
phenomenon highlights a functional difference between machine and human vision, which relies more on shape information~\cite{geirhos_imagenet-trained_2019}. Texture bias has been linked to models' inability to handle corruptions and out of distribution samples, and has been hypothesized to contribute to adversarial vulnerability~\cite{chen_shape_2022, geirhos_generalisation_2020, geirhos_imagenet-trained_2019}. 

However, existing approaches for measuring and mitigating texture bias have not yet been able to capture how naturally occurring textures impact model robustness in real-world settings. Existing works have relied on the texture-shape cue conflict dataset~\cite{geirhos_imagenet-trained_2019}, which contains images with object silhouettes of one object class (e.g., outline of a cat) superimposed on the texture of another object class (e.g., elephant skin). While effective at understanding a model's overall tendency toward one feature or the other, this approach has multiple limitations: (1) the texture used in these samples is predetermined and hand-selected to match specific classes, preventing the discovery of unexpected associations between textures and objects, (2) there is an overwhelming amount of texture present in the images, potentially increasing the models' preference towards texture compared to natural settings and (3) the quantification of whether a model is texture biased is solely based on the models' prediction on this artificially constructed data, leaving the role of texture bias in real-world classifications and the influence of naturally occurring textures largely unexplored.

\begin{figure}[t]
\centering
\includegraphics[width=\linewidth]{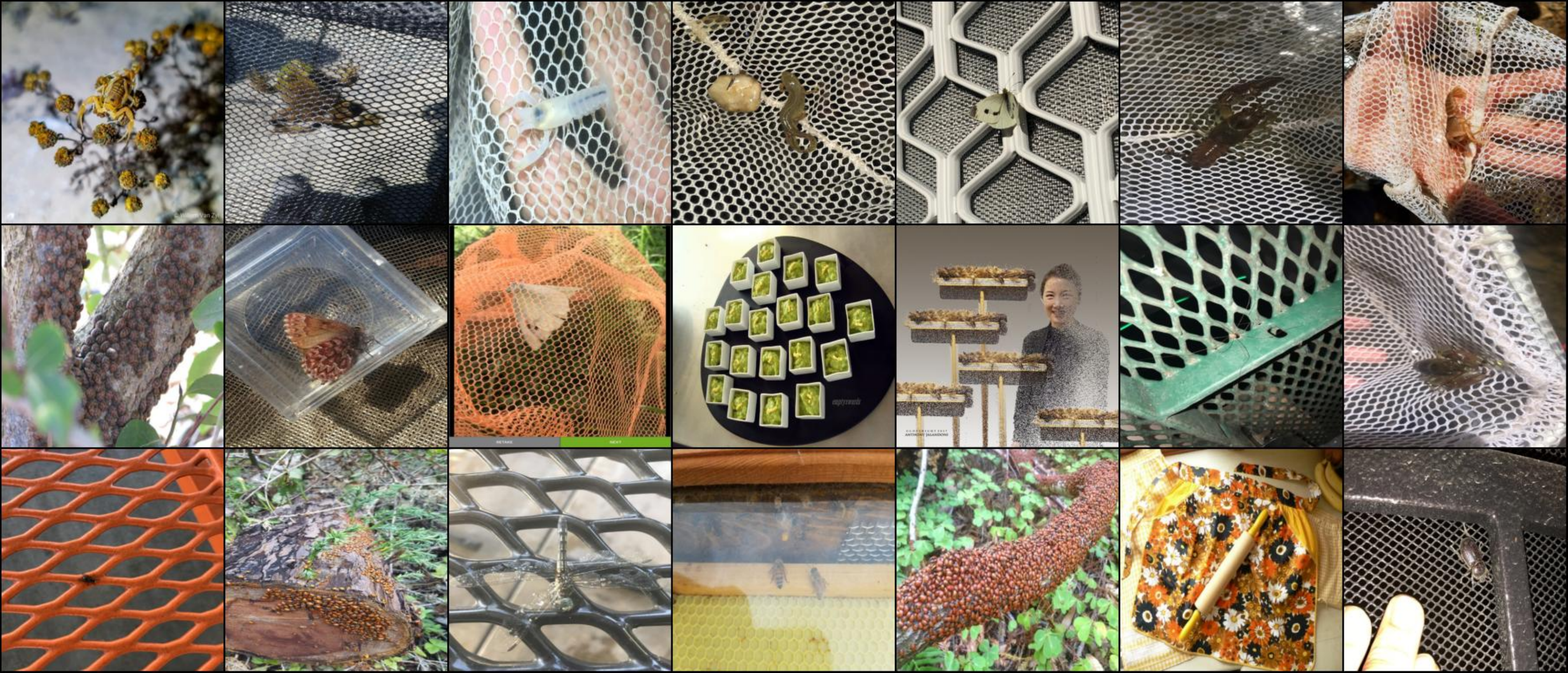}
    \caption{ImageNet-A~\cite{hendrycks_natural_2021} examples misclassified as honeycombs on ResNet50.}
    \label{fig:honeycomb_classifications}
\end{figure}

We hypothesize that textures serve as a primary signal for driving classification on real data. This hypothesis was inspired by the observation that ``natural adversarial examples''~\cite{hendrycks_natural_2021}--- natural samples that cause confident yet incorrect predictions--- are often heavily textured. Additionally, when visualizing misclassified samples, we found that those assigned to the same incorrect class share extremely similar textures, despite being unrelated to the actual object (example shown in \autoref{fig:honeycomb_classifications}), suggesting that these misclassifications may be due to the presence of specific textures. 

In this paper, we introduce a novel approach to evaluate models' bias toward texture. Central to this approach is the \tavfull{} (\tav{}), a new metric that leverages diverse texture data to quantify the associations between textures and objects. We first compute \tav{} using the Prompted Textures Dataset (PTD)~\cite{hoak_synthetic_2024}, a modern corpus of texture images. 
Leveraging the \tav{}, we identify textures present in images by comparing similarity in model responses on real data to those on texture images, and thus study how these textures influence model classification.

We investigate how textures influence model accuracy and robustness through a three-stage evaluation. First, we analyze the properties of the \tav{} metric to assess how models respond to textures in isolation, which informs whether textures alone can drive confident model predictions. Next, we evaluate how naturally occurring textures influence model predictions on real images. Here, we compare model accuracy and confidence when classifying images that contain textures frequently associated with the object class versus images with less-frequently occurring textures. Finally, we study how bias towards texture impacts robustness, where we analyze how the textures present in natural adversarial examples lead to confident mispredictions.

Our findings demonstrate that model classifications are heavily driven by the presence of specific textures, impacting both accuracy and robustness. Despite the fact that models are not purposefully trained to recognize textures, we find that models highly confidently predict isolated textures---we observe over 25,000 texture images were classified as objects with over 96\% confidence. On ImageNet validation data, we find that models are highly reliant on specific textures they learned during training. Comparing performance on images that contained the dominant texture for an object class with images that contained other textures, we find that models exhibit up to a 66\% difference in accuracy and 40\% difference in confidence. Finally, we provide strong evidence that the existence of natural adversarial examples is due to misaligned textures---we find that over 90\% of these samples contain textures that are not dominant for the object class of their true label.

In summary, this work provides a comprehensive investigation into how textures influence model performance in real-world settings. By introducing the \tav{} metric and applying it to real data, we offer a novel approach to identifying textures in images and analyzing their role in model decision-making. Our findings demonstrate that texture bias plays a critical role in both model accuracy and robustness, especially in challenging scenarios like natural adversarial examples. This approach offers new insights into how texture bias influences real data classifications and opens new avenues for assessing and addressing  model trustworthiness through a new lens. We release our code and data at \codeurl{}.

\section{Background}\label{sec:background}

\subsection{Texture Bias}

Geirhos et al.~\cite{geirhos_imagenet-trained_2019} first uncovered the existence of texture bias in CNNs. In this work, they introduced the texture-shape cue conflict dataset, which consists of images across 16 different object classes containing the texture of one object with the shape of another object (e.g., elephant skin texture on the shape of a cat). With this dataset, they found that humans would classify images more often in line with the ``shape class'' (e.g., cat from the previous example) while CNNs would classify them as their ``texture class'' (e.g., an elephant from the previous example). This intriguing and groundbreaking finding identified a major high-level functional difference between human and machine vision. 

Hoak and McDaniel~\cite{hoak_explorations_2024} introduced the notion of texture learning, which focuses on the identification of textures learned by object classification models. Rather than quantify how biased models are towards texture, they describe how to construct texture-object associations, which quantifies the relationship between textures and objects. To compute these texture-object associations, they analyze how frequently texture images from the Describable Textures Dataset (DTD)~\cite{cimpoi_describing_2014} are classified as different objects. They find that models learn both ``expected'' textures (e.g., a waffled texture for a waffle iron object) as well as ``unexpected'' textures (e.g., a polka-dotted texture for a shower curtain object). They additionally find that these unexpected associations can reveal information about bias in training data, highlighting the importance of studying texture bias beyond hand-selected textures.

Brendel and Bethge~\cite{brendel_approximating_2019} introduced the concept of BagNet, a neural network architecture that operates solely on local image patches. BagNets were designed to study whether CNNs could make accurate predictions using only texture-like information from small patches of an image. Their findings confirmed that CNNs could indeed classify images with high accuracy using only local information (e.g., textures), further highlighting the dominance of texture in CNN decision-making processes. Here, they investigate if textures are sufficient for classification, while we investigate if textures are necessary for classification. This conclusion also agrees with prior works, which discuss how textures are simpler for CNNs to learn, and how these models may take shortcuts in their learning to only generalize based on the easiest features to learn~\cite{geirhos_shortcut_2020}.

\subsection{Natural Adversarial Examples}

The ImageNet-A dataset~\cite{hendrycks_natural_2021} contains images coined as ``Natural Adversarial Examples.'' Adversarial examples, first shown in Deep Neural Networks in 2014~\cite{szegedy_intriguing_2014}, are inputs designed to induce model misclassification. They are crafted by adding specially produced, human imperceptible perturbations which are designed to cause mispredictions through any number of attack methods~\cite{biggio_evasion_2013, goodfellow_explaining_2015, madry_towards_2019, moosavi-dezfooli_deepfool_2016, sheatsley_space_2022, carlini_towards_2017, papernot_limitations_2015}. Such adversarial examples are intriguing in that models often classify such inputs with alarming confidence, even though the underlying semantics of the image have been clearly preserved. Natural Adversarial Examples are conceptually similar to adversarial examples in that such inputs are also confidentially misclassified by models, except that the ``perturbation'' applied to induce misclassification was not explicitly crafted by an adversary, but instead exists in natural settings. 

In this paper, we hypothesize that the texture bias present in object recognition models represents a sufficient condition for the existence of natural adversarial examples. In other words, natural adversarial examples likely contain a spurious texture strongly associated with other object classes that models are sensitive to due to their inherent texture bias.

\section{Methodology}
\subsection{Texture-Object Associations}\label{sec:text_obj}

Prior findings and evaluations on texture bias have been limited to the shape-texture cue conflict dataset~\cite{geirhos_imagenet-trained_2019}, which: (1) only analyzes 16 different object classes, (2) contain textures that are selected and labeled based on what textures are assumed to be associated with certain objects and (3) does not investigate how this texture bias translates to impact on real data (e.g., images not in the texture-shape cue conflict dataset) classifications.

In this work, we introduce the \tavfull{} (\tav{}), a metric that quantifies the relationship between textures and the object classes a model predicts through texture object associations~\cite{hoak_explorations_2024}. We construct these associations by analyzing model predictions on diverse texture data, which allows us to scale up our texture bias evaluation and discover (rather than assume) what textures are learned by models when classifying certain objects. Furthermore, \tav{} represents how models interpret different kinds of textures, which we later leverage by comparing how models interpret real data images with naturally occurring textures, enabling our texture bias evaluation on real data. 

Simply put, the \tav{} captures how much a model relies on specific textures to make predictions about objects. It assigns a score to each texture-object pair, where a higher score indicates a stronger association between the texture and the object class predicted by the model. For example, a high \tav{} value between striped textures and zebra objects means that models strongly learned to look for the presence of stripes when classifying zebras. The top 50 strongest object-texture associations can be found in \autoref{fig:text_obj_bar_pairs} and are discussed later in \autoref{sec:texture_response}. 

To construct this metric, we first use synthetic texture images from the Prompted Textures Dataset (PTD)~\cite{hoak_synthetic_2024} as input to the model and observe the model's predictions. We then compute how frequently these textures are classified as certain objects, creating a matrix that records the associations between textures and objects. The TAV then incorporates several factors, such as how likely a texture is to be classified as a specific object and how concentrated these classifications are across all object classes, forming a product of probabilities and entropies. The final result is a matrix that is $n \times m$, where $n$ is the number of texture classes and $m$ is the number of object classes, and each value in the matrix contains the association score between a given texture object class pair. Below, we detail the exact mathematical formulation of the TAV and the properties it captures.

\subsubsection{Constructing the \tav{}}
Let $D_t$ denote the portion of the dataset
containing images of texture class $t$, $f_\theta$ as the trained object
classification model, $x$ as an image from the selected portion of the
dataset, and $c$ is the object class of interest (one of the object classes from
the trained model). We first construct our metric by using the texture images as input to the model and getting their predictions argmax$(f_\theta (x))$. We then record how many times samples from each texture class were classified as each object class:

\begin{equation}
  N_{ij} = \sum_{x \in D_i} \boldsymbol{\mathbbm{1}}(\text{argmax}(f_\theta (x)) = j)
\end{equation}

Here, $i$ represents an index into a texture class, $j$ is an index into an object class, and $D_i$ is the subset of the dataset that contains all samples of texture class $i$. With these counts we then form the basis for our \tavfull{} (\tav{}) metric. In constructing a metric that accurately captures the association between textures and objects, we have a few key desired properties for texture-object pairs that are strongly associated. First, samples of a particular texture should have high probability of being classified as an object class. The probability of texture class $i$ being predicted as object class $j$ is represented by:

\[
PT_{ij} = \frac{N_{ij}}{\sum_{j} N_{ij}}
\]

At the same time, we also want to ensure that, if a texture class and object class are strongly associated then, out of all the samples that were predicted as belonging to that object class, a large majority of those samples should belong to the provided texture class. The probability of a prediction on object class $j$ being from a sample belonging to texture class $i$ is:

\[
PO_{ij} = \frac{N_{ij}}{\sum_{i} N_{ij}} 
\]

Additionally, we want the object classes that these texture samples are being predicted as to have predictions that are concentrated to a few texture classes, otherwise, this would suggest that the object class didn't learn an over-reliance on textures present in the training data. In other words, objects that are associated with \textit{many} textures are not strongly associated with \textit{any} textures. To capture the concentration, we take the complement of the entropy (one minus the entropy) of the object class predictions. The entropy of object class $j$ is:

\[
OH_j = -\sum_{i} \left( PO_{ij} \log PO_{ij} \right) 
\]

Finally, texture classes should also be concentrated to a few object classes, because if they weren't, then that would suggest that classifying the texture class is akin to randomly guessing and thus is not a significant or interesting texture for the model. To account for this, we measure the concentration of the prediction distributions for each texture class through the complement of the entropy (one minus the entropy) of the distribution. The entropy of texture class $i$ is represented as:

\[
TH_i = -\sum_{j} \left( PT_{ij} \log PT_{ij} \right)
\]

Putting each of these components together, the \tavfull{} (\tav{}) is shown in \autoref{eq:strength_of_association}. Higher \tav{} for a given texture object pair ($i$, $j$) corresponds to stronger associativity between the two.

\begin{equation}
\text{TAV}_{ij} = PT_{ij} \cdot (1 - TH_i) \cdot PO_{ij} \cdot (1 - OH_j) 
\label{eq:strength_of_association}
\end{equation}

\begin{figure}
    \centering
    \includegraphics[width=\linewidth]{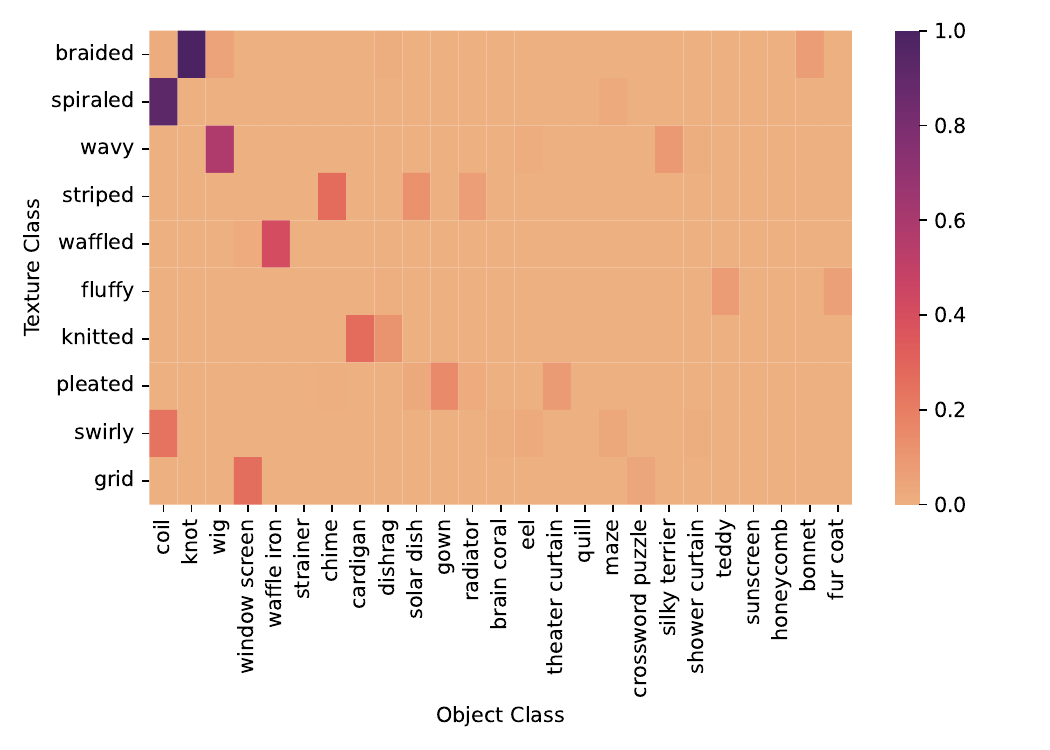}
    \caption{A subset of the \tav{} matrix.}
    \label{fig:tav_demo}
\end{figure}

With the \tav{}, we now have a direct relationship between textures and model object classes in the form of a matrix that is $n \times m$, where $n$ is the number of texture classes and $m$ is the number of object classes. \autoref{fig:tav_demo} shows shows a demonstration of the \tav{} computed on the Prompted Textures Dataset. For space, the entire \tav{} cannot be displayed. This figure contains 10 texture classes (out of 56 total) and 25 object classes (out of 1000 total). This matrix provides us with two key properties. 

First, by examining individual elements in the \tav{} (e.g., for one texture class and object class pair) we have a measure of how associated the two are, providing us with a good estimation of how strongly that texture was learned during training to identify that object class. In \autoref{sec:texture_response} we further explore this first property and analyze the associations we find in individual elements in the \tav{}.

Second, by examining entire rows in the \tav{}---where a row corresponds to a given texture class, and is a vector of $m$ (object classes) length---we have a good estimation of how a model will predict texture images of that texture class (i.e., roughly what the output probabilities would be from the model if given an image of that texture class). Next, we detail how we leverage these distributions to extend our study to real data.

\subsection{Identifying Textures Present in Images}\label{sec:tid}

To understand how naturally occurring textures influence real data classifications, we must first be able to identify textures present in images in order to then analyze their influence on models. To address this, we develop a texture identification method that builds upon the \tav{}, which maps texture classes to the object classification distributions they produce. This mapping provides a comprehensive view of how models associate specific textures with object classes based on their responses to texture images.

Our goal is to extend this analysis to real data by comparing how models respond to both texture data and real images. We hypothesize that if models exhibit similar behavior when classifying real images and texture images, it suggests that the real image contains the corresponding texture. Thus, by measuring the similarity between the model's output probabilities for a real image and those of a texture class from the \tav{} matrix, we can infer which texture is present in the real image.

To formalize this process, we introduce the Texture Identification (TID), which assigns a texture class to each real image. The TID works by comparing the model's softmax outputs for a real image to each row in the \tav{} matrix (representing each texture) and selecting the texture with the highest similarity to the image's output distribution. More formally, the \textit{TID} for an image is calculated as:

\begin{equation}
  \text{TID}(x) = \argmax_{i} \frac{\text{softmax}(f_\theta(x)) \cdot \text{TAV}_{i}}{\|\text{softmax}(f_\theta(x))\| \cdot \|\text{TAV}_{i}\|}
  \label{eq:tid}
\end{equation}

\begin{figure}[t]
  \centering
  \includegraphics[width=\linewidth]{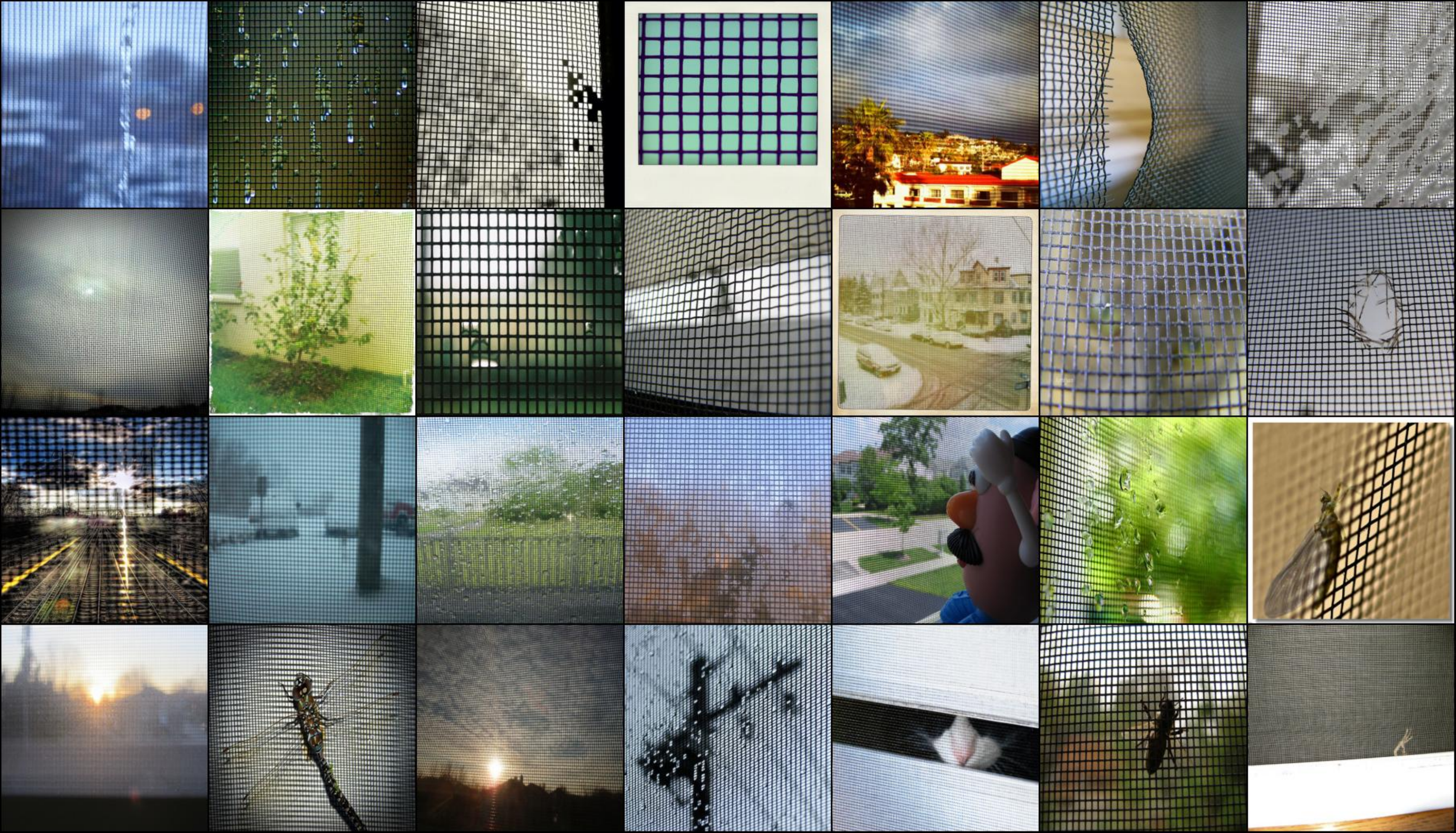}
  \caption{Images from the ImageNet validation set identified as having grid textures.}
  \label{fig:grid_examples}
\end{figure}

With the TID, we assign a texture class to each of the images in both the ImageNet validation set and ImageNet-A set, which we further detail in \autoref{sec:results}. In \autoref{fig:grid_examples} we show a subset of the samples from the ImageNet validation set that we identified as having the ``grid'' texture through the TID, demonstrating how well this technique captures the textures that are visually present in the images. More examples can be found in \autoref{appendix:tid_examples}. From this, we can see that the TID identifies textures that are well aligned with what we would expect (i.e., the images in the figure all have a grid pattern). We next perform a more comprehensive evaluation of efficacy of this approach through a human evaluation. 

\subsubsection{Validation of TID}\label{sec:human_eval}

To evaluate the accuracy of our TID metric, we conduct a human evaluation on texture identification and compare it to the results of the TID. We measure the accuracy of the TID by calculating the ratio of images where human evaluators identify the same texture as the TID.

To conduct this study, we employ 10 graduate students in computer science. Each participant is given a set of 200 images, each with 4 texture options. For each image, participants record which texture best matches the textures present in the image. The images are selected from the ImageNet validation set, and the 4 total texture options include the 1 texture identified by the TID plus 3 randomly selected textures. For detailed instructions and questions provided to participants, see Appendix \ref{appendix:human_eval}.

\begin{figure*}[t]
    \centering
    \includegraphics[width=0.8\linewidth]{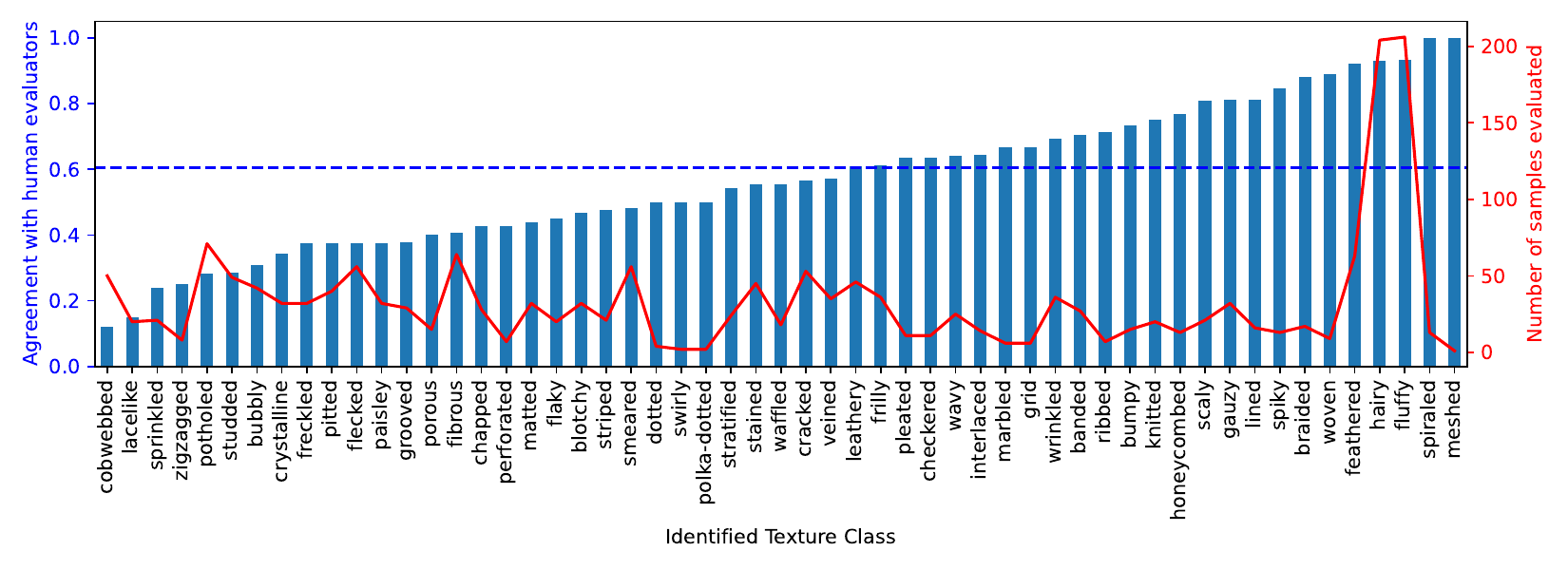}
    \caption{Average agreement with human evaluators and number of samples evaluated for each predicted texture class. Horizontal line shows the overall agreement with human evaluators.}
    \label{fig:human_eval}
\end{figure*}

In \autoref{fig:human_eval}, we show the agreement rate between the textures identified by human evaluators and the textures identified through the TID---which represents the percentage of samples where human evaluators identified the same texture as the TID--on each of the texture classes and overall across all samples. We observe that consistency between human evaluators and the TID is highly dependent on the texture class. For instance, texture classes such as “hairy” and “fluffy” show a strong agreement, with human evaluators matching the TID over 90\% of the time. In contrast, for more ambiguous textures like “cobwebbed,” the agreement rate drops significantly, nearing random chance at 13\%. This discrepancy may be due to (a) the subtlety of certain textures in the dataset, which may not provide enough prominent examples, and (b) human evaluators being better attuned to certain textures, resulting in varying prediction rates across texture classes.

\begin{figure}
    \centering
    \includegraphics[width=\linewidth]{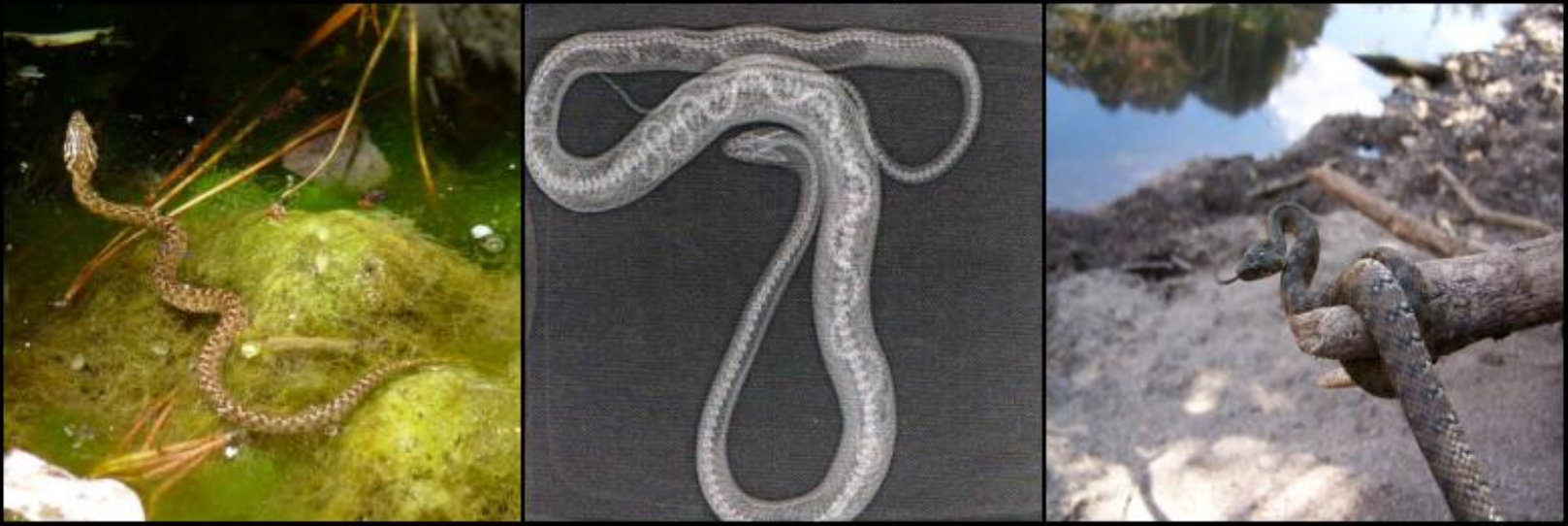}
    \caption{Samples labeled as having a ``swirly'' texture by human evaluators and a ``flecked'' texture by the TID.}
    \label{fig:human_eval_examples}
\end{figure}

One key challenge we identified is that human vision is inherently shape-biased, as noted by Geirhos et al.~\cite{geirhos_imagenet-trained_2019}. This bias means humans might overlook finer texture details, especially when textures are intertwined with other visual cues like shape. For example, as shown in \autoref{fig:human_eval_examples}, several images of snakes were identified by human evaluators as having a “swirly” texture based on the coiled shape of the snake’s body. However, the TID identified the texture as “flecked,” focusing on the intricate patterning of the snake’s scales. This divergence illustrates that, unlike models, human evaluators might prioritize the overall shape and contour of an object over the specific surface texture.

Moreover, when multiple textures are present in an image, human evaluators may select the texture related to the central object, while the TID may be more sensitive to background textures or patterns across the entire image. This phenomenon adds complexity to interpreting texture identification, especially in real-world settings where images often contain multiple overlapping textures.

Despite these challenges, our evaluation finds that the TID aligns with human evaluators in 61\% of cases. This result is well above the 25\% baseline for random guessing, and given the inherent difficulties in human texture perception—especially in scenarios where textures are subtle or co-occur with other cues—this agreement rate demonstrates the effectiveness of the TID. Identifying textures in real data remains a nuanced task, and while there is room for improvement, the TID offers a promising method for texture identification that complements human perception in challenging cases.

The TID provides us with a powerful tool, and enables us to automatically and accurately identify textures in real images based on model responses, allowing for a more detailed analysis of texture bias in uncontrolled, natural settings. This capability is crucial for evaluating texture bias on real data, and enables our investigation on the influence textures have on model accuracy and robustness.

\section{Results}\label{sec:results}
In this work, we hypothesize that texture presence heavily influences model accuracy, confidence, and robustness. Towards this, we investigate the following research questions:
\begin{enumerate}
    \item \textit{How do models respond to texture alone?}
    % \item \textit{How does our texture identification compare to human evaluators?}
    \item \textit{Do textures drive classification in real images?}
    \item \textit{Can texture bias explain the existence of natural adversarial examples?}
\end{enumerate}
\subsection{Setup}
\subsubsection{Experimental Details} All models used in our experiments are pretrained on ImageNet~\cite{russakovsky_imagenet_2015} and obtained from torchvision~\cite{marcel_torchvision_2010} with the default model weights. The model was evaluated on two datasets using the following data preprocessing steps: (1) resize the image to 256$\times$256, (2) center crop the image to 224$\times$224, (3) normalize the image using the mean and standard deviation of the ImageNet training dataset. All experiments were run across 12 NVIDIA A100 GPUs. Complete code to replicate experiments can be found at \codeurl{}.

For consistency and brevity, all results reported in this section are on the ResNet50~\cite{he_deep_2015} model. For completeness, we additionally evaluated the following models: ResNet18, ResNet152, EfficientNetB0~\cite{tan_efficientnet_2020}, DenseNet121~\cite{huang_densely_2018}, DenseNet169, Inception-v3~\cite{szegedy_rethinking_2015}, and ConvNeXt~\cite{liu_convnet_2022}. These models were chosen to validate our results on a wide variety of architectures, model sizes, and on CNN-VIT hybrids. Extended results on all models can be found in the corresponding appendix sections. We found the results across all models to be highly consistent with the ResNet50 results presented here.

\subsubsection{Datasets} Here, we describe how we initialize the \tav{} with texture data, and how the subsequent \tav{} matrix is used to identify textures present in images.

\noindent\textbf{Prompted Textures Dataset.} The Prompted Textures Dataset (PTD)~\cite{hoak_synthetic_2024} is a dataset of high-resolution textures. The dataset contains 362,880 images spanning 56 texture classes. Images of textures within the dataset have dimensions equal to 256x256, enabling the dataset to be readily usable by a variety of popular pre-trained ImageNet models. We use the Prompted Textures Dataset to calculate the \tav{}, which describes the association between textures and objects. Moreover, the dataset contains a variety of textures, thereby eliminating assumptions on what kinds of textures models should be biased towards, as discussed in \autoref{sec:background}.

\noindent\textbf{ImageNet.} ImageNet~\cite{russakovsky_imagenet_2015} is a large-scale, high-resolution image dataset designed for object recognition. The dataset contains 1,000 object classes with 1,281,167 training images, 50,000 validation images, and 100,000 test images. Images within the dataset are preprocessed to have dimensions equal to 256x256. Given the popularity of ImageNet as the canonical benchmark for object classification models and the high resolution of the images compared to other popular image datasets (e.g., CIFAR 10 or 100), we use it to assess the degree to which the textures present within the images bias model predictions.

\noindent\textbf{ImageNet-A.} ImageNet-A~\cite{hendrycks_natural_2021} is a hand-curated set of ImageNet-like samples that ImageNet models are confidently incorrect classifying. The dataset contains 7,500 (confidently mispredicted) images across the 200 selected object classes sourced from Flickr and iNaturalist. Like ImageNet, images are also preprocessed to have dimensions equal to 256x256. We use ImageNet-A to evaluate our hypothesis that natural adversarial examples contain textures strongly associated with specific object classes that cause misclassification. In this way, our analysis of natural adversarial examples provides further evidence that texture represents a sufficient condition to drive model predictions.

\subsection{Models' Response to Textures}\label{sec:texture_response}

Using our new \tavfull{} (\tav{}) metric, we can now measure the strength of
texture associativity for each object class. To do this, we ran every texture image (from the PTD)
through a variety of pretrained object classification models, giving us an object class prediction (and a confidence in that prediction) for each texture class.

\begin{figure}
  \centering
  \includegraphics[width=\linewidth]{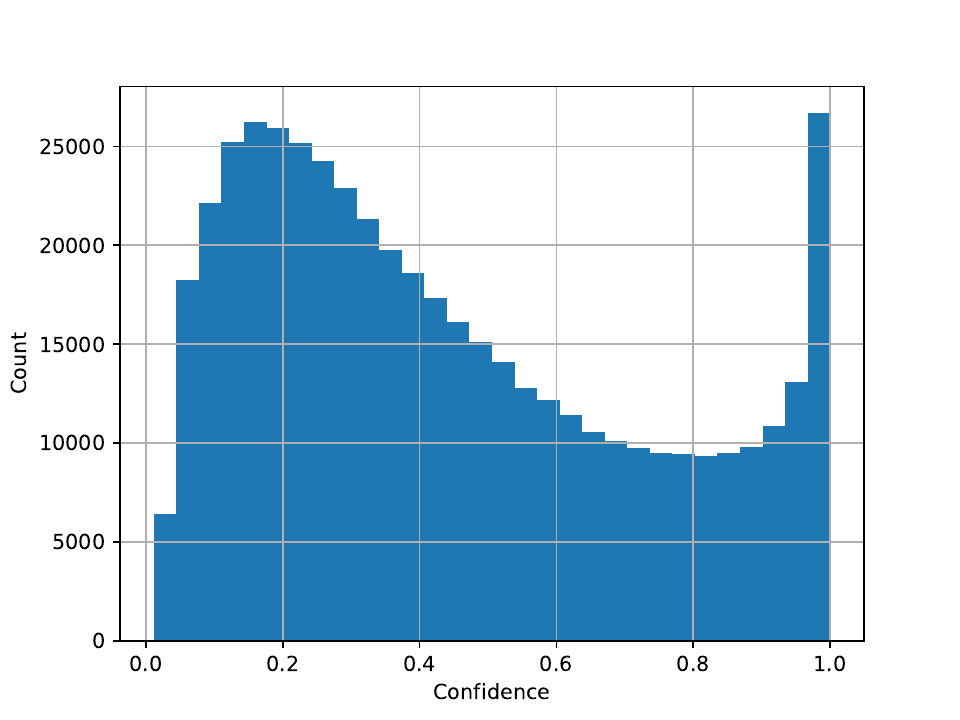}
  \caption{Confidence histogram of the classification of texture images on ResNet50.}
  \label{fig:confidence_hist}
\end{figure}

In addition to using the class prediction to compute the \tav{} (discussed later in this section), we also analyze the confidence the model had in making the prediction, which characterizes how strongly models respond to texture images.
\autoref{fig:confidence_hist} shows a histogram of confidence values for all the texture images in PTD.
Interestingly, we found that these confidence values were very often \textit{far} above a
random guessing rate (0.001 for 1000 classes) even despite the fact that these texture images are unrelated to the data the model was trained on (ImageNet). Additionally, over 25,000 samples were at or close to 100\% confidence. This further demonstrates the
prevalence of texture learning/bias, given how responsive models are to textured
images that are not even from the same distribution as their training or test
data. Results on additional models can be found in \autoref{appendix:confidence_hist}.

\begin{figure*}[t]
  \centering
  \includegraphics[width=\textwidth]{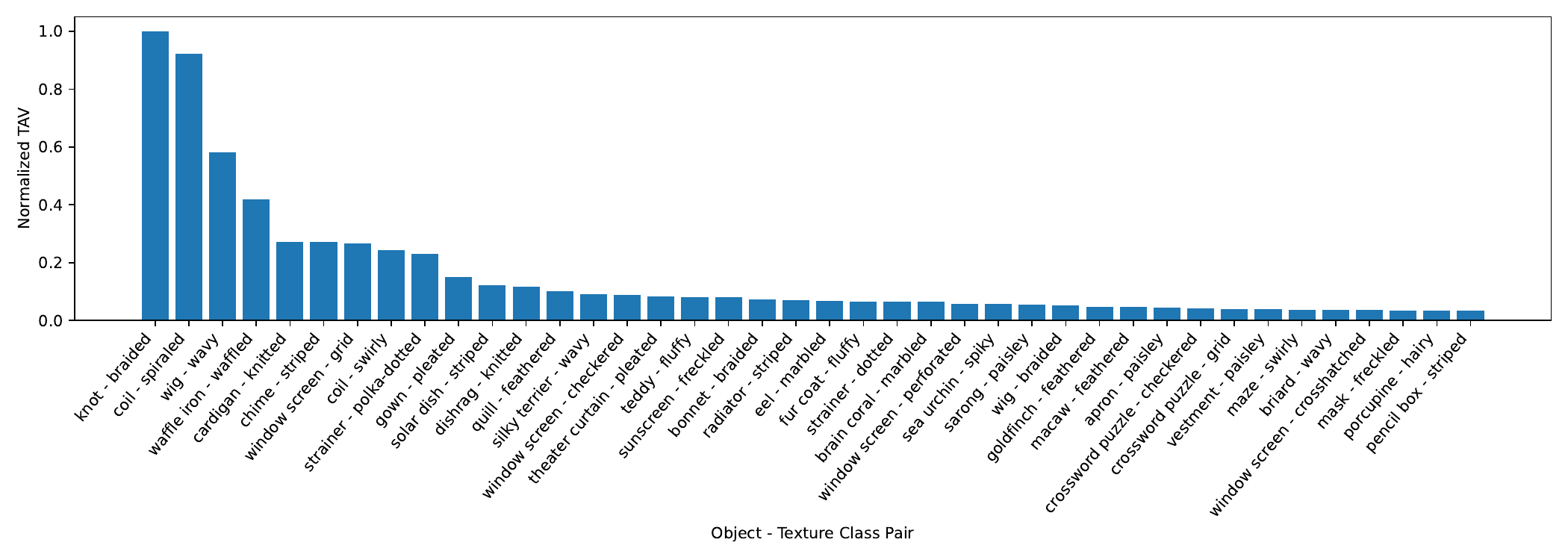}
  \caption{Top 50 object-texture class pairs with the highest \tav{} values on
  ResNet50.}
  \label{fig:text_obj_bar_pairs}
\end{figure*}

We now use the class predictions of all the texture images to calculate the
\tav{} for every texture-object class pair for each model. This resulted in
56,000 \tav{} (1,000 object classes $\times$ 56 texture classes) values for each
model. In \autoref{fig:text_obj_bar_pairs} we show the object-texture class
pairs with the 50 highest \tav{} values on ResNet50 (other models
can be found in Appendix \ref{appendix}).

From this figure, we see a variety of texture-object relationships uncovered.
Notably, despite the fact that these texture images are out of distribution from
the training data of these models (ImageNet), the models are still able to form
strong associations between textures and objects. This strongly suggests that the models
are heavily learning from and relying on textures to classify images, supporting the results of prior work~\cite{geirhos_imagenet-trained_2019,brendel_approximating_2019, hoak_explorations_2024}. Further,
the associations that are uncovered are often intuitive. In other words, grids being
classified as window screens or waffled textures being classified as waffle
irons \textit{makes sense}. This property highlights that our
methodology can readily and accurately capture the kinds of textures that models
learn in various object classes.

\noindent\textbf{Takeaway}: Models \textit{confidently} predict texture images, even when they are not explicitly trained to, demonstrating that texture alone is sufficient for confident classification.

\subsection{Texture Bias on Real Images}\label{sec:tb_real}
With confirmation of our texture identification method introduced in \autoref{sec:tid}, we now have the capacity to investigate how naturally present textures in object images impact model classification.

% In this section, we investigate the core hypothesis of this work: that textures
% are the prominent driver of object classification tasks. 
Previously, the construction of \tav{} and investigation on how models respond to textures was done with respect to the Prompted Textures Dataset (PTD). Here, we now leverage the \tav{} and texture identification technique to study texture bias in real images (e.g., naturally occurring, in-distribution, clean validation data).
We begin this investigation by identifying trends in the textures present in
clean images from the ImageNet validation set and whether the texture present in the image impacts the
classification of that image. If textures can be varied in an image without
changing the classification, the model is likely not primarily driven by
texture. However, if changing the texture in an image changes the
classification, then texture has a large influence on the model. 

\begin{figure*}
    \centering
    \includegraphics[width=\textwidth]{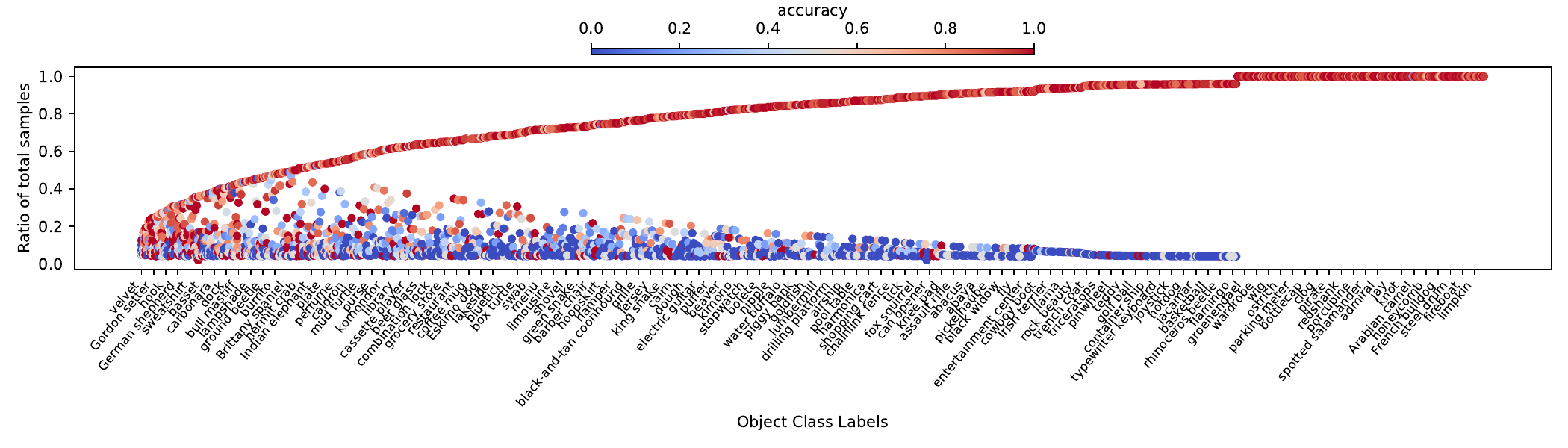}
    \caption{Scatter plot of the texture groupings present in each label by how many samples are in each group (normalized by number of samples in each label). The color of the points represents the accuracy of the model on the samples in that group.}
    \label{fig:texture_object_count_labels_accuracy}
\end{figure*}

We first investigate if textures present in images impact model accuracy. We begin by identifying the texture present in every sample of the ImageNet validation set. In
\autoref{fig:texture_object_count_labels_accuracy} we separate out the images in
the ImageNet
validation set by their true labels (1000 object classes). For each of the 1000
labels, we group images together based on the texture they contain (as identified by the TID). Each point in the plot represents a texture class that was present in at least one of the samples for the corresponding object label on the x axis. The y axis shows how
many samples contain that texture (normalized by the total number of samples belonging to the object label). We sort the ordering of the object class labels by the ratio of the total number of samples belonging to the most frequently occurring texture class (i.e., their largest y value). The object labels at the rightmost part of the plot had all their samples containing one texture, because there is only one point for each label, and the ratio of total samples of that point is 100\%, meaning that 100\% of the samples contained that texture (and thus 0\% of the samples contained a different texture). We then color each point based on the average accuracy of the samples that reside within that point (i.e., the samples that belong to that label and contain that texture). Due to the large number of object labels (1000 total), not all label names could be displayed on the axis. A subset of label names are shown on the axis, but the points corresponding to all 1000 object labels are present in the plot.

From this figure we observe 2
interesting trends: (1) there is a striking separation between the accuracy on
the most dominant (most frequently occurring) textures and the least dominant (least frequently occurring) textures, \textbf{showing that
the presence of the most heavily learned texture for a given object class is
the deciding factor in the accurate classification of that object}, and (2) many of the
object classes even have only a single texture present in their images, suggesting that
despite any other variation in the images of that object class, texture still
serves as a meaningful and accurate signal for the model to classify images. 

To provide a more consolidated view of the results discussed here, we also analyze these trends in aggregate across labels and on a variety of models. \autoref{tab:corr_accuracy} displays the correlation between the ratio of total samples containing a texture class and the accuracy of the model on those samples (e.g., the correlation between the y axis and color of \autoref{fig:texture_object_count_labels_accuracy}), as well as the average number of texture classes in samples of an object class label (e.g., the average number of points per object label in \autoref{fig:texture_object_count_labels_accuracy}). The high correlation further demonstrates that accuracy is heavily influenced by texture presence. Interestingly, we also find that the average number of textures found in an object class label fluctuates with the model. Particularly, within model classes such as the ResNets and DenseNets, models tend to have a lower number of textures they are associated with as they get larger. The largest model, ConvNext, also has the smallest average number of textures out of all models. Overall, this could suggest one of two things: either the larger models tend to be less biased towards texture, because they learn to rely on fewer textures, or the larger models are more biased toward texture, because they tend to strongly associate with few, specific textures. We investigate this further in \autoref{sec:tb_natadv}.

\begin{table}[t]
    \centering
    \caption{Label statistics across models.}
    \begin{tabular}{l|rr}
\toprule
Model & Accuracy correlation & Avg. \# of textures\\
\midrule
convnext-base & 0.68 & 2.59 \\
densenet121 & 0.63 & 3.60 \\
densenet169 & 0.66 & 3.39\\
efficientnet-b0 & 0.57& 3.42 \\
inception-v3 & 0.68 & 3.24 \\
resnet152 & 0.64 & 2.84 \\
resnet18 & 0.61 & 4.13 \\
resnet50 & 0.63 & 3.42 \\
\bottomrule
\end{tabular}
    \label{tab:corr_accuracy}
\end{table}

\begin{figure}
    \centering
    \includegraphics[width=\linewidth]{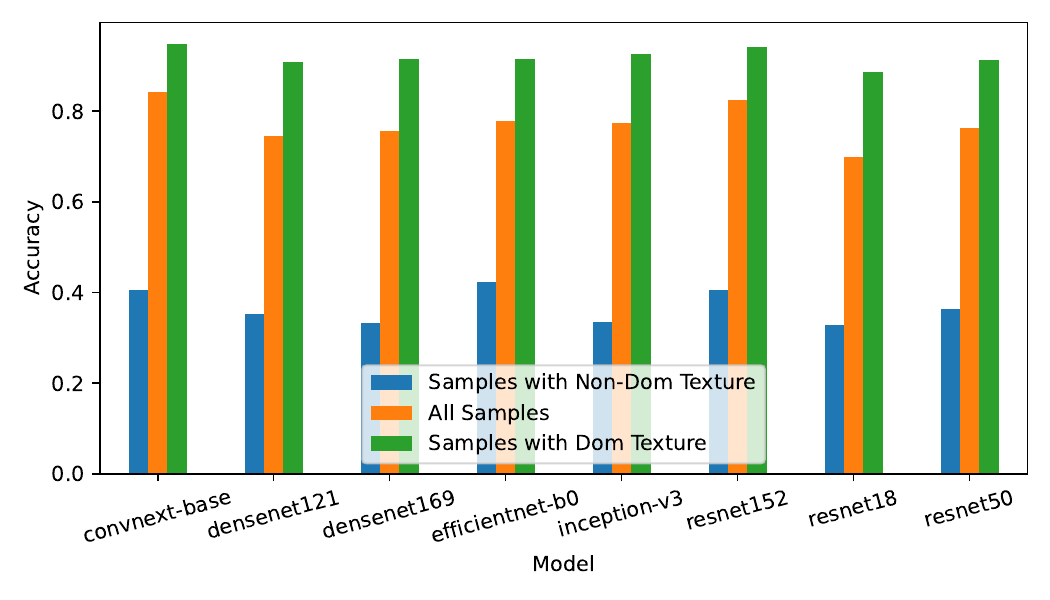}
    \caption{The accuracy of samples that do and do not contain the dominant texture class for their label, along with the model accuracy on all samples regardless of texture, across models.}
    \label{fig:bar_acc_texture_ratio}
\end{figure}

Finally, \autoref{fig:bar_acc_texture_ratio} shows the accuracy of the samples that contain the dominant texture (most frequently occurring) for their label class, samples that contain a non-dominant texture for their label class, and baseline model accuracy across all samples. These results demonstrate that models are up to 67\% more accurate on samples containing a dominant texture than they are on samples containing a non-dominant texture. Even across a wide variety of model architectures, models are consistently and vastly more accurate on samples that contain dominant textures over the non-dominant textures, supporting that texture presence is largely responsible for accurate classification. 

% \todo{talk about how there are some classes that learn multiple texture representations and how we can see this from the classes that are spread. however, from this we also see that the models are still most confident on their dominant texture than they are on any secondary textures. maybe also analyze what those textures are, some of them could be highly matched}

\begin{figure*}
\centering
\includegraphics[width=\textwidth]{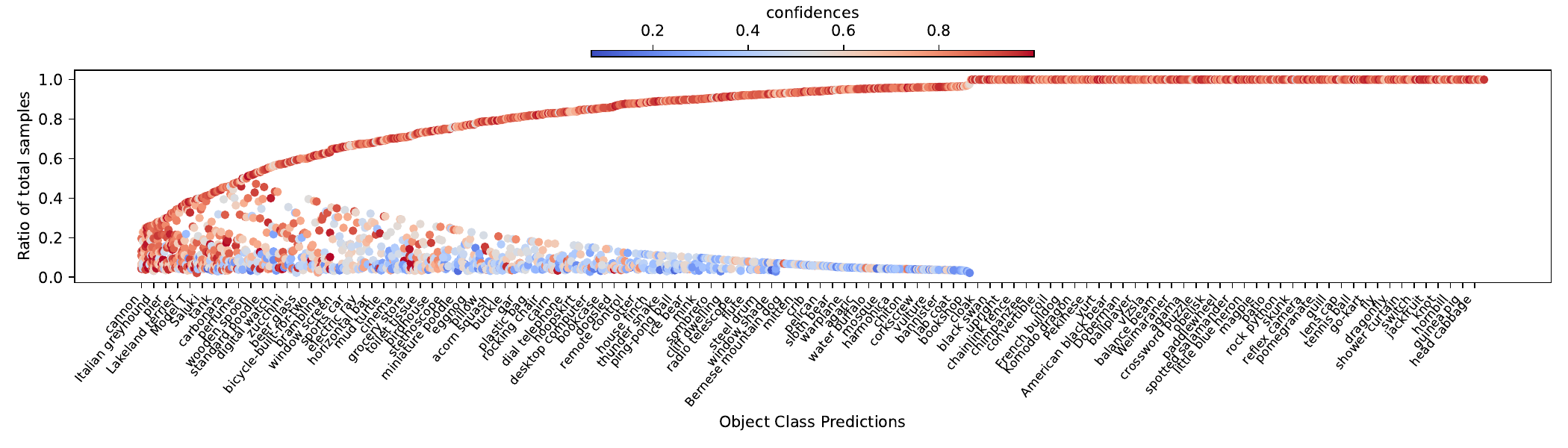}
    \caption{Scatter plot of the texture groupings present in each object prediction by how many samples are in each group (normalized by total number of samples per object prediction class). The color of the points represents the average confidence of the model on the samples in that group.}
    \label{fig:texture_object_count_preds_conf}
\end{figure*}

We now want to investigate how texture presence impacts model confidence. In
\autoref{fig:texture_object_count_preds_conf} we perform a similar analysis to
the previous accuracy analysis but instead of grouping by true labels, we group by the
model's predictions and color by model confidence rather than model accuracy. The specific procedure is as follows: we begin by identifying the texture present in every sample of the ImageNet validation set. We then separate out the images in
the ImageNet
validation set by the object class they are \textit{predicted} as, not labeled as, totaling 1000 object classes. For each of the 1000 prediction object classes, we group images together based on the texture they contain (as identified by the TID). Each point in the plot represents a texture class that was present in at least one of the samples for the corresponding prediction class on the x axis. The y axis shows how
many samples contain that texture (normalized by the total number of samples that were predicted as each object class). We sort the ordering of the object class predictions by the ratio of the total number of samples belonging to the most frequently occurring texture class (i.e., their largest y value). The object prediction classes at the rightmost part of the plot had all their samples containing one texture, because there is only one point for each prediction class, and the count of that point is 100\%, meaning that 100\% of the samples contained that texture (and thus 0\% of the samples contained a different texture). We then color each point based on the average \textit{confidence} (rather than accuracy) of the samples that reside within that point (i.e., the samples that were predicted as that object class and contain that texture). Due to the large number of object classes (1000 total), not all object class names could be displayed on the axis. The points corresponding to all 1000 object labels are present in the plot, but only a subset of prediction class names are shown on the axis.

This figure shows a similar trend to our findings on model accuracy; the model is more confident in its
predictions when the image contains the most dominant texture for that object
class. This suggests that containing a dominant texture for a given object class
is necessary for the model to make a confident prediction. Thus, supporting our
hypothesis that conflicting texture could lead to confidently wrong predictions
as long as the dominant texture for a non-true label class is present.

\autoref{tab:corr_conf} displays the correlation between the ratio of total samples containing a texture class and the average confidence of the model on those samples, as well as the average number of texture classes in samples of an object class prediction. The high correlation further demonstrates that confidence is heavily influenced by texture presence. Similarly to \autoref{tab:corr_accuracy}, we also find that smaller models tend to have a lower number of textures per class.

\begin{table}[t]
    \centering
    \caption{Prediction statistics across models.}
    \begin{tabular}{l|rr}
\toprule
Model & Confidence correlation & Avg. \# of textures\\
\midrule
convnext-base & 0.64 & 1.98\\
densenet121 & 0.64 & 2.62 \\
densenet169 & 0.65 & 2.42 \\
efficientnet-b0 & 0.56 & 2.68 \\
inception-v3 & 0.67 & 2.35 \\
resnet152 & 0.59 & 2.10 \\
resnet18 & 0.63 & 3.08 \\
resnet50 & 0.60 & 2.53 \\
\bottomrule
\end{tabular}
    \label{tab:corr_conf}
\end{table}

\begin{figure}
    \centering
    \includegraphics[width=\linewidth]{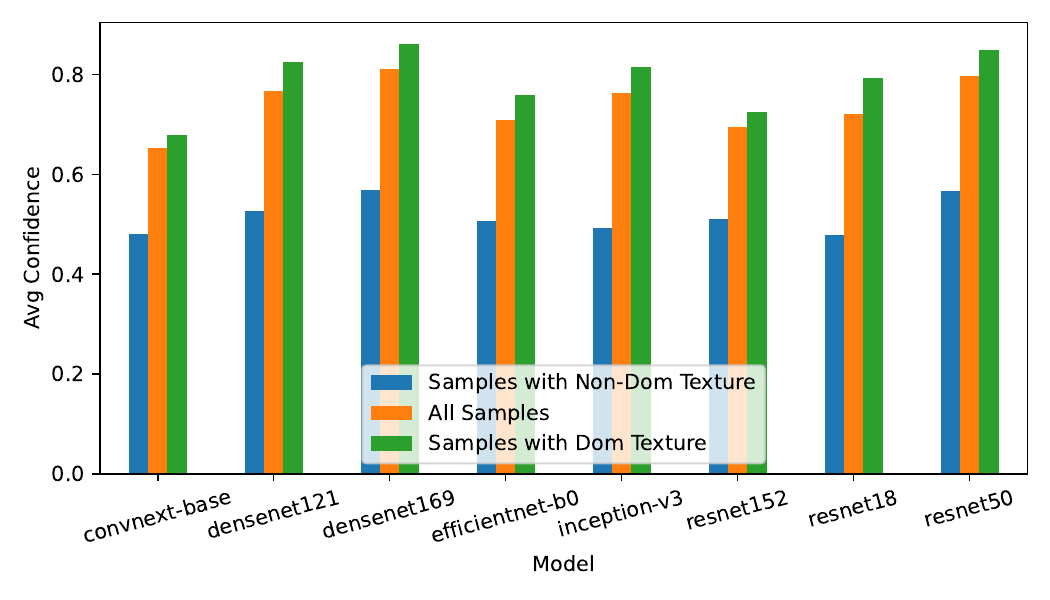}
    \caption{The average confidence of samples that do and do not contain the dominant texture class for their object prediction class, along with the average confidence on all samples regardless of texture, across models.}
    \label{fig:bar_conf_texture_ratio}
\end{figure}

We analyze \autoref{fig:bar_conf_texture_ratio} in the same way as \autoref{fig:bar_acc_texture_ratio}; here we display the average model confidence on samples containing the dominant texture and non-dominant textures for each sample's prediction class. Again, we can see that across all models, confidence is the highest when the dominant texture is present in the image. Across all models, we observed a difference of up to 40\% model confidence on the samples with versus without the dominant texture for the prediction class.

\noindent\textbf{Takeaway:} Confident, accurate classifications \textit{necessitate} the presence of textures associated with the corresponding object class.

\subsection{Texture Bias in Natural Adversarial Examples}\label{sec:tb_natadv}

Natural adversarial examples~\cite{hendrycks_natural_2021} are samples that are confidently misclassified (similar to adversarial examples) but these examples occur naturally within clean data. Lying somewhere between an adversarial example and simple error, natural adversarial examples provide us with data that allows deeper investigation into the kinds of errors models make. 

Based on the key results from the previous section, we were interested if textures could be used to explain inaccurate and confident predictions. Here we hypothesize that the existence of natural adversarial examples is due to the presence of a conflicting texture in the image. As we saw in the last section, the presence of particular textures can determine the confidence in a model's prediction. This suggests that any differences in texture from the dominant texture of the true label can skew predictions.

We begin investigating this hypothesis by gathering model predictions on the ImageNet-A dataset (accuracy on ImageNet-A for each model can be found in \autoref{tab:imneta_acc}) and identifying how frequently the texture present in the image aligns with the most dominant texture for the object class of both the prediction and the label. Here, we analyze three different textures: (1) using the object class that each natural adversarial example is predicted as, we get the \textit{prediction texture} by identifying the most dominant texture from the ImageNet data (from the upper envelope of \autoref{fig:texture_object_count_preds_conf}) for that object class, (2) using the object class that each natural adversarial example is labeled as, we get the \textit{label texture} in the same way, and (3) using the natural adversarial example image, we identify the texture present in the image according to the TID. We say that there is agreement on the predictions if the texture found in the image is the same as the prediction texture. Similarly, there is label agreement if the texture found in the image matches the label texture. 

\begin{figure}
    \centering
    \includegraphics[width=\linewidth]{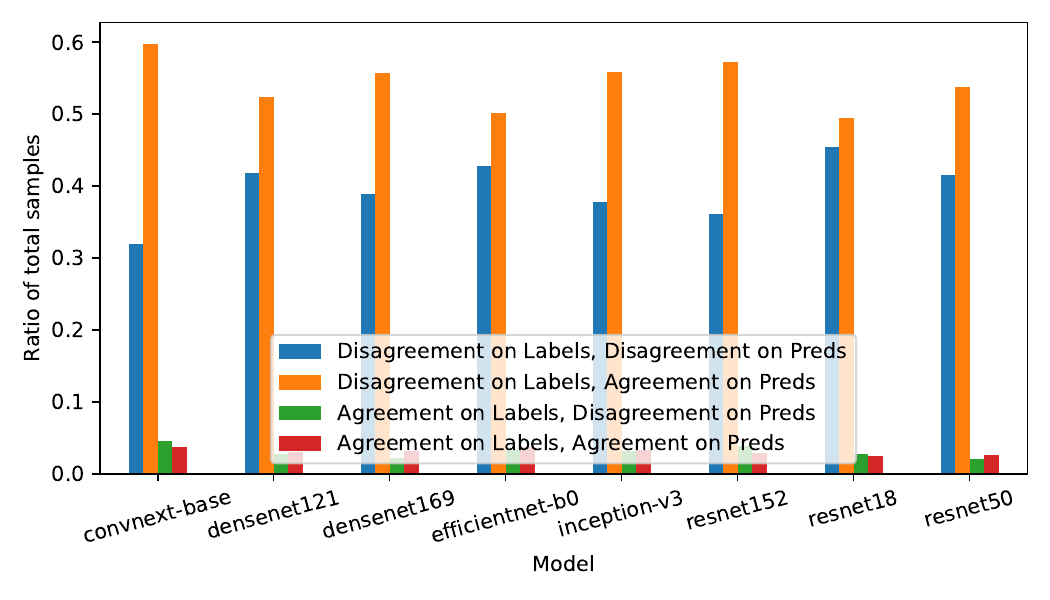}
    \caption{The average alignment between the identified and the most common texture for a sample's object prediction and label class on ImageNet-A.}
    \label{fig:texture_alignment_rates}
\end{figure}

In \autoref{fig:texture_alignment_rates} we show the ratio of total samples in the ImageNet-A dataset that contain a texture that agree with their prediction texture, label texture, neither, and both. From this figure, we can see that the texture present in samples very rarely has agreement with the texture corresponding to its label. In up to 60\% of samples, the texture present in the image matches the most common texture (i.e., the dominant texture) for the prediction class \textit{and} does not match the dominant texture for the label class. More than 90\% of the samples contain textures that disagree with the texture associated with their true label (i.e., samples in the blue and orange bars).

Interestingly, there are very few samples where the texture in the image is the same as the texture in \textit{both} the prediction and the label object class. We work with a total of 56 different texture classes and 1000 different object classes, meaning that there are roughly 18 object classes that are mapped to each texture class. For many misclassifications where the label and prediction are very close (e.g., great white sharks and tiger sharks), we would expect that the texture class for these two object classes would be the same despite the fact that the sample is still being misclassified, resulting in a more ``understandable'' error (where it is easy to see why the model made a mistake). 

Contrary to these ``understandable'' errors, we see that the samples of natural adversarial examples represent a class of errors that goes beyond this, as the samples typically contain a texture that is completely different from its label texture. \textbf{We find that the presence of this different, misaligned texture explains natural adversarial examples' confident mispredictions.}

% \todo{This makes sense because we know it isn't changing the object, so it must be changing the texture.}

\begin{figure*}
    \centering
    \includegraphics[width=\linewidth]{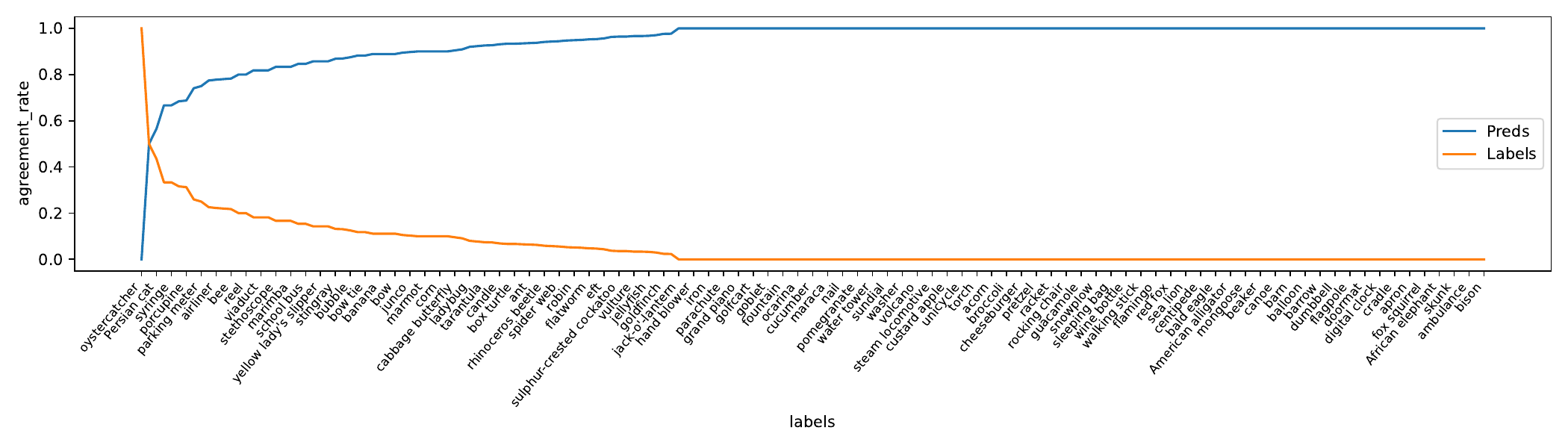}
    \caption{The rate of agreement between the textures identified in ImageNet-A images and the textures predominantly found in the respective object labels and predictions for those images, separated by the object label class.}
    \label{fig:agreement_rates}
\end{figure*}

Next, we further study the cases where label and prediction texture alignment disagree, and investigate only the samples that differ in their agreement with prediction and label textures (i.e., the orange and green bars of \autoref{fig:texture_alignment_rates}). In \autoref{fig:agreement_rates} we show the rate of agreement between the textures identified in ImageNet-A images and the textures predominantly found in the respective labels and predictions for those images, separated by the label class. We find that: (1) over 99\% of object class labels have more samples that align with their prediction texture than the label texture and (2) over 60\% of class labels have 100\% alignment across their samples with their prediction texture, and 0\% alignment with their label texture. 

We further investigated the single class (``oystercatcher'') that had more alignment with the label texture than the prediction texture and found that this class only contained a single sample, the model had relatively low confidence (12\%) when classifying this sample, and the textures for the prediction and label classes were, respectively, ``potholed'' and ``grooved'' which are conceptually similar textures.

From this, we can see that the natural adversarial examples are highly aligned with the texture of their prediction, and highly misaligned with the texture of their true label, explaining their confident misprediction. 

Finally, we investigate the question \textit{are natural adversarial examples more textured than clean data?} For this analysis, we use the TID, but rather than selecting a texture, we look at the magitude of the similarity between textures and object images (i.e., \autoref{eq:tid} but with \textit{max} rather than \textit{argmax}). 

\begin{figure}
    \centering
    \includegraphics[width=\linewidth]{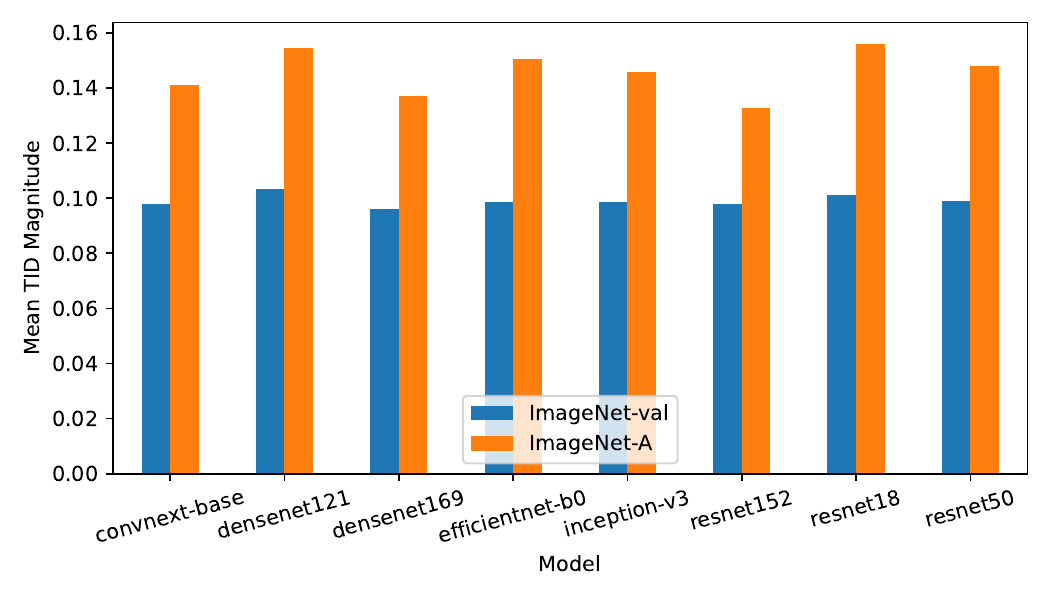}
    \caption{Mean TID magnitude for ImageNet validation data and ImageNet-A.}
    \label{fig:mean_tid_mag}
\end{figure}

In \autoref{fig:mean_tid_mag} we show the mean TID magnitude of both the ImageNet validation set and the ImageNet-A dataset. From this, we can see that images from ImageNet-A are consistently more similar to texture images, supporting that natural adversarial examples are more textured than clean validation data.

\noindent\textbf{Takeaway:} Natural adversarial examples are a consequence of texture bias, and their confident yet incorrect predictions can be explained by the fact that they contain textures that are not aligned with their true label.

\section{Related Work}
\noindent\textbf{Identifying textures and texture bias.} Geirhos et al.~\cite{geirhos_imagenet-trained_2019} introduced one of the first works that investigated model bias towards texture. In addition to creating the first benchmark to measure texture bias, they also found that models were capable of learning shapes alone by altering the training data to destroy local texture information.

Hermann et al.~\cite{hermann_origins_2020} set out to uncover what properties or training schemes lead to increased texture bias. They found that random crops used in data augmentation during training were the most likely to lead to more texture biased models. These results suggest that texture bias may not be due to the model alone, but also due to the data that the model sees. 

% While prior work has investigated measuring, mitigating, and explaining the
% existence of texture bias~\cite{geirhos_imagenet-trained_2019,
% geirhos_partial_2021, hermann_origins_2020, gatys_texture_2015}, there has been
% limited research done on how to effectively evaluate what kinds of textures
% models are biased towards, and how such biases impact the trustworthiness of the
% model. 
Recent work~\cite{hoak_explorations_2024} has introduced the notion
of texture learning, which studies the extent to which a model learns and relies
on textures for classification. While this new approach to uncovering learned
textures is promising, results have still been limited to smaller scale
datasets and investigates texture bias at a class level rather than a sample specific level.

Interpretability frameworks such as Network Dissection~\cite{bau_network_2017} serve as useful tools to aid in making models more interpretable and can aid in highlighting learned textures by visualizing concepts learned by certain object classes. However, the textures it can identify are based on the Describable Textures Dataset (DTD) \cite{cimpoi_describing_2014}, the same smaller scale dataset used in \cite{hoak_explorations_2024}.

To the best of our knowledge, there has not yet been a method that is capable of analyzing texture bias on real data classifications. Thus, comparison to existing methods is challenging due to differences in evaluation capabilities.

\noindent\textbf{Reducing texture bias.} One of the most prevalent works that aims to mitigate the effect of texture bias is the same work that introduced the concept of texture bias~\cite{geirhos_imagenet-trained_2019}. From the observation that models will often classify images according to their texture rather than their shape, the authors train shape biased models by taking the ImageNet training set and distorting the texture signals in the images by utilizing style transfer~\cite{gatys_texture_2015, gatys_image_2016} to inpaint various artistic textures into the images. Training the same architectures on this new ImageNet dataset, called Stylized ImageNet~\cite{geirhos_imagenet-trained_2019}, they find that the resulting models are not only more shape biased, but also more accurate and robust to common corruptions (i.e., ImageNet-C~\cite{hendrycks_benchmarking_2019}). Similar approaches have also been introduced using other methods to distort texture information in training data, such as SDbOA~\cite{he_shift_2023}.

In addition to works aiming to reduce texture bias, other works argue that both texture and shape serve as important cues for image classification models, and that models focusing exclusively on one cue or the other will lead to undesirable errors~\cite{li_shape-texture_2021}. To achieve a balanced model, the authors introduce a shape-texture debiased training scheme wherein models are trained on images with conflicting shape and texture, similar to the texture-shape cue conflict dataset~\cite{geirhos_imagenet-trained_2019}. The authors find that training these debiased models leads to better accuracy and robustness on both ImageNet-A~\cite{hendrycks_natural_2021} and ImageNet-C~\cite{hendrycks_benchmarking_2019}. 

It has also been shown that adversarial training~\cite{goodfellow_explaining_2015, huang_learning_2016, kurakin_adversarial_2017, madry_towards_2019}--- the process of training machine learning models on adversarial examples, rather than clean ones, for the purposes of boosting robustness to test time adversarial examples---can result in models that are more biased towards shape rather than texture~\cite{zhang_interpreting_2019, chen_shape_2022}. However, these models also tend to have lower clean accuracy, making them less desireable for use in non-adversarial settings.
\section{Discussion}

\noindent\textbf{When is texture bias undesirable?} An ideal model, and one that functions similarly to the human visual system, will rely on a more balanced ratio of both texture and shape information when classifying objects. Currently, we see that models are more biased towards texture than they should be, but there has yet to be any comprehensive studies on what situations warrant learning texture versus those that don't. For example, in order to learn how to classify a waffle, models may necessarily have to rely on the presence of a waffled texture, as this texture serves as the primary signal that differentiates a waffle from a pancake. However, when classifying aprons, it may not be necessary for models to learn to look for a paisley pattern\footnote{Both the waffle object to waffled texture and apron object to paisley texture examples were selected based on actual associations we observed to be among the strongest that the model learned, shown in \autoref{fig:text_obj_bar_pairs}.}, since the presence of this pattern or not does not change the (human) classification of this object. We find working towards characterizing when textures should be learned or not to be a very important and interesting direction for future work, which will hopefully be further enabled with the techniques introduced in this paper.

While there have been some approaches that propose ways to mitigate texture bias, these findings have been with respect to prior texture bias benchmarks, which analyzes texture bias on synthetic data rather than real, naturally occurring data as we do in this work. Furthermore, many of these approaches set out to make models as shape-biased as possible (e.g., by destroying texture information from training data~\cite{geirhos_imagenet-trained_2019}) such that models are only able to rely on shape and are constructed using heavy data augmentation. We believe that future works on mitigating texture bias should be with respect to a balanced view of texture and shape, focus on potential for both model-driven and data-driven methods, and target specific instances where texture bias may be undesirable. We believe that such models will benefit from increased accuracy and robustness, as well as lower texture bias.

\noindent\textbf{Texture identification.} In this work, we leverage synthetic texture data to construct the \tav{}, which serves as an estimation for how models respond to and predict textures of various classes. We then identify textures present in real images by comparing the output probabilities on these images to rows in the \tav{}. We designed our methodology this way for two key reasons. First, we leverage the PTD because it serves as a good source of \textit{labeled} texture data (i.e., we know what kind of texture is present in the image). Other approaches that work towards extracting textures from images, such as style transfer~\cite{gatys_neural_2015, gatys_image_2016}, were not used in this work because (1) the textures that are extracted must come from source images, which require additional considerations when selecting and (2) since these methods extract information at multiple intermediate layers in models, the technique must be adapted for each model, imposing variation in the quality of textures extracted. Similarly, techniques such as patch cross-correlation, patch mean variance, or frequency analysis, can provide a measure of the "textureness" of an image, but lack the identification of what the texture is. Second, we identify textures present in images by comparing model outputs on real data to rows in the \tav{}. Prior works have most commonly identified textures through texture classification models, which are typically CNNs trained on texture images~\cite{simon_deep_2020}. We opted for the former approach because it (1) does not require additional model training, as we are solely operating on pre-trained object classification models, which also eliminates additional sources of bias that may be imposed by introducing more models and (2) rather than focusing on gathering the most accurate texture classifications possible, we focus on characterizing how models interpret different textures, which is more relevant to the goals we set out to achieve when researching textures that models are biased towards.

\noindent\textbf{Limitations and future work.} As described above, we designed our methodology specifically to fit within the goals we wanted to accomplish with this study. However, no method is without limitations. Specifically, by using texture data as part of our metric, we necessarily rely on the texture data we have available to us, and thus could be missing textures that models learn whose structure is not present in existing texture datasets. While leveraging texture data has several desirable properties, we also recognize that different methods may be more appropriate for different evaluations. Additionally, since our method operates solely on model outputs and not on any hidden representations, we do not characterize how texture bias may evolve throughout layers in the model. To address both limitations, we see a wide variety of exciting directions for future work on integrating interpretability techniques to help identify textures learned by models.

We also find expanding this evaluation to analyze bias of other elements of imagery to be a worthwhile topic for future work. Prior works have demonstrated that CNNs may also be overly biased towards color~\cite{rafegas_color_2018} and that color-based alterations of images can lead to successful adversarial examples~\cite{chen_explore_2020, kantipudi_color_2020}. With an appropriate color dataset in place of the PTD, future work could adapt a similar methodology to the one introduced here to construct associations between colors and objects and identify the impact of color bias on model accuracy and robustness.

In this work we found that the existence of natural adversarial examples can be explained by texture bias, and that the confident mispredictions of these samples arise from the fact that they contain textures that are misaligned with their true label. While this is an important result for understanding confident mispredictions, there are many other kinds of adversarial data. In future work, we plan to explore how universal adversarial examples~\cite{moosavi-dezfooli_universal_2017} and traditional adversarial examples crafted with various attack methods~\cite{biggio_evasion_2013, goodfellow_explaining_2015, madry_towards_2019, moosavi-dezfooli_deepfool_2016, sheatsley_space_2022, carlini_towards_2017, papernot_limitations_2015} may also be explained by texture bias. Additionally, we also plan to investigate how texture bias extends to other models, such as larger vision transformers, defenses such as adversarial training, and other tasks such as object detection.

\section{Conclusions}
In this work, we introduced the \tavfull{} (\tav{}), a novel metric for quantifying the extent to which models rely on textures when classifying objects. Our findings reveal that texture bias is a significant factor influencing model robustness and accuracy on real data. We demonstrated that natural adversarial examples can be attributed to texture bias, with a majority of such examples arising from the presence of textures that are misaligned with samples' true labels, leading to confident mispredictions. By providing a deeper understanding of how textures drive model behavior, our approach offers a new pathway for assessing and mitigating texture-driven vulnerabilities in machine learning systems. In the future, we aim to explore how other aspects of trustworthy machine learning, such as fairness and interpretability, as well as other facets of robustness, like adversarial examples, might also be influenced or explained by texture bias.
% Future work will explore methodologies for further reducing texture bias and improving model robustness, particularly in the face of complex, real-world data distributions.
% \input{0_notes_and_outlines}

\section{Acknowledgments}

This material is based upon work supported by, or in part by, the National Science Foundation under Grant No. CNS2343611. Any opinions, findings, and conclusions or recommendations expressed in this publication are those of the author(s) and do not necessarily reflect the views of the National Science Foundation, or the U.S. Government. The U.S. Government is authorized to reproduce and distribute reprints for government purposes notwithstanding any copyright notation hereon.

\printbibliography

@misc{chen_shape_2022,
	title = {The shape and simplicity biases of adversarially robust {ImageNet}-trained {CNNs}},
	url = {http://arxiv.org/abs/2006.09373},
	doi = {10.48550/arXiv.2006.09373},
	publisher = {arXiv},
	author = {Chen, Peijie and Agarwal, Chirag and Nguyen, Anh},
	month = sep,
	year = {2022},
}

@article{simon_deep_2020,
	series = {Third {International} {Conference} on {Computing} and {Network} {Communications} ({CoCoNet}'19)},
	title = {Deep {Learning} based {Feature} {Extraction} for {Texture} {Classification}},
	volume = {171},
	issn = {1877-0509},
	url = {https://www.sciencedirect.com/science/article/pii/S1877050920311613},
	doi = {10.1016/j.procs.2020.04.180},
	journal = {Procedia Computer Science},
	author = {Simon, Philomina and V, Uma},
	month = jan,
	year = {2020},
	pages = {1680--1687},
}

@misc{gatys_neural_2015,
	title = {A {Neural} {Algorithm} of {Artistic} {Style}},
	url = {http://arxiv.org/abs/1508.06576},
	language = {en},
	publisher = {arXiv},
	author = {Gatys, Leon A. and Ecker, Alexander S. and Bethge, Matthias},
	month = sep,
	year = {2015},
}

@misc{huang_learning_2016,
	title = {Learning with a {Strong} {Adversary}},
	url = {http://arxiv.org/abs/1511.03034},
	doi = {10.48550/arXiv.1511.03034},
	publisher = {arXiv},
	author = {Huang, Ruitong and Xu, Bing and Schuurmans, Dale and Szepesvari, Csaba},
	month = jan,
	year = {2016},
}

@inproceedings{he_shift_2023,
	address = {Paris, France},
	title = {Shift from {Texture}-bias to {Shape}-bias: {Edge} {Deformation}-based {Augmentation} for {Robust} {Object} {Recognition}},
	copyright = {https://doi.org/10.15223/policy-029},
	shorttitle = {Shift from {Texture}-bias to {Shape}-bias},
	url = {https://ieeexplore.ieee.org/document/10377235/},
	doi = {10.1109/ICCV51070.2023.00147},
	language = {en},
	booktitle = {2023 {IEEE}/{CVF} {International} {Conference} on {Computer} {Vision} ({ICCV})},
	publisher = {IEEE},
	author = {He, Xilin and Lin, Qinliang and Luo, Cheng and Xie, Weicheng and Song, Siyang and Liu, Feng and Shen, Linlin},
	month = oct,
	year = {2023},
	pages = {1526--1535},
}

@inproceedings{gatys_image_2016,
	address = {Las Vegas, NV, USA},
	title = {Image {Style} {Transfer} {Using} {Convolutional} {Neural} {Networks}},
	isbn = {978-1-4673-8851-1},
	url = {http://ieeexplore.ieee.org/document/7780634/},
	doi = {10.1109/CVPR.2016.265},
	language = {en},
	booktitle = {2016 {IEEE} {Conference} on {Computer} {Vision} and {Pattern} {Recognition} ({IEEE/CVF} {Conference} {on} {Computer} {Vision} {and} {Pattern} {Recognition} {(CVPR)})},
	publisher = {IEEE},
	author = {Gatys, Leon A. and Ecker, Alexander S. and Bethge, Matthias},
	month = jun,
	year = {2016},
	pages = {2414--2423},
}

@inproceedings{marcel_torchvision_2010,
	address = {New York, NY, USA},
	series = {{MM} '10},
	title = {Torchvision the machine-vision package of torch},
	isbn = {978-1-60558-933-6},
	url = {https://dl.acm.org/doi/10.1145/1873951.1874254},
	doi = {10.1145/1873951.1874254},
	booktitle = {Proceedings of the 18th {ACM} international conference on {Multimedia}},
	publisher = {Association for Computing Machinery},
	author = {Marcel, Sébastien and Rodriguez, Yann},
	month = oct,
	year = {2010},
	pages = {1485--1488},
}

@article{geirhos_shortcut_2020,
	title = {Shortcut {Learning} in {Deep} {Neural} {Networks}},
	volume = {2},
	issn = {2522-5839},
	url = {http://arxiv.org/abs/2004.07780},
	doi = {10.1038/s42256-020-00257-z},
	number = {11},
	journal = {Nature Machine Intelligence},
	author = {Geirhos, Robert and Jacobsen, Jörn-Henrik and Michaelis, Claudio and Zemel, Richard and Brendel, Wieland and Bethge, Matthias and Wichmann, Felix A.},
	month = nov,
	year = {2020},
	pages = {665--673},
}

@inproceedings{cimpoi_describing_2014,
	address = {Columbus, OH, USA},
	title = {Describing {Textures} in the {Wild}},
	isbn = {978-1-4799-5118-5},
	url = {https://ieeexplore.ieee.org/document/6909856},
	doi = {10.1109/CVPR.2014.461},
	language = {en},
	booktitle = {2014 {IEEE} {Conference} on {Computer} {Vision} and {Pattern} {Recognition}},
	publisher = {IEEE},
	author = {Cimpoi, Mircea and Maji, Subhransu and Kokkinos, Iasonas and Mohamed, Sammy and Vedaldi, Andrea},
	month = jun,
	year = {2014},
	pages = {3606--3613},
}

@inproceedings{chen_explore_2020,
	address = {New York, NY, USA},
	series = {{CODASPY} '20},
	title = {Explore the {Transformation} {Space} for {Adversarial} {Images}},
	isbn = {978-1-4503-7107-0},
	url = {https://dl.acm.org/doi/10.1145/3374664.3375728},
	doi = {10.1145/3374664.3375728},
	booktitle = {Proceedings of the {Tenth} {ACM} {Conference} on {Data} and {Application} {Security} and {Privacy}},
	publisher = {Association for Computing Machinery},
	author = {Chen, Jiyu and Wang, David and Chen, Hao},
	month = mar,
	year = {2020},
	pages = {109--120},
}

@article{ballester_performance_2016,
	title = {On the {Performance} of {GoogLeNet} and {AlexNet} {Applied} to {Sketches}},
	volume = {30},
	copyright = {Copyright (c)},
	issn = {2374-3468},
	url = {https://ojs.aaai.org/index.php/AAAI/article/view/10171},
	doi = {10.1609/aaai.v30i1.10171},
	language = {en},
	number = {1},
	journal = {Proceedings of the AAAI Conference on Artificial Intelligence},
	author = {Ballester, Pedro and Araujo, Ricardo},
	month = feb,
	year = {2016},
	note = {Number: 1},
}

@misc{hoak_synthetic_2024,
	title = {On {Synthetic} {Texture} {Datasets}: {Challenges}, {Creation}, and {Curation}},
	shorttitle = {On {Synthetic} {Texture} {Datasets}},
	url = {http://arxiv.org/abs/2409.10297},
	doi = {10.48550/arXiv.2409.10297},
	publisher = {arXiv},
	author = {Hoak, Blaine and McDaniel, Patrick},
	month = sep,
	year = {2024},
}

@inproceedings{moosavi-dezfooli_universal_2017,
	title = {Universal adversarial perturbations},
	url = {http://arxiv.org/abs/1610.08401},
	doi = {10.48550/arXiv.1610.08401},
	booktitle = {{IEEE/CVF} {Conference} {on} {Computer} {Vision} {and} {Pattern} {Recognition} {(CVPR)} 2017},
	publisher = {arXiv},
	author = {Moosavi-Dezfooli, Seyed-Mohsen and Fawzi, Alhussein and Fawzi, Omar and Frossard, Pascal},
	month = mar,
	year = {2017},
}

@inproceedings{kantipudi_color_2020,
	title = {Color {Channel} {Perturbation} {Attacks} for {Fooling} {Convolutional} {Neural} {Networks} and {A} {Defense} {Against} {Such} {Attacks}},
	url = {http://arxiv.org/abs/2012.14456},
	doi = {10.48550/arXiv.2012.14456},
	booktitle = {{IEEE} {Transactions} on {Artificial} {Intelligence}},
	publisher = {arXiv},
	author = {Kantipudi, Jayendra and Dubey, Shiv Ram and Chakraborty, Soumendu},
	month = dec,
	year = {2020},
}

@inproceedings{zhang_interpreting_2019,
	title = {Interpreting {Adversarially} {Trained} {Convolutional} {Neural} {Networks}},
	url = {http://arxiv.org/abs/1905.09797},
	language = {en},
	booktitle = {{International} {Conference} {on} {Machine} {Learning} {(ICML)} 2019},
	publisher = {arXiv},
	author = {Zhang, Tianyuan and Zhu, Zhanxing},
	month = may,
	year = {2019},
}

@inproceedings{kurakin_adversarial_2017,
	title = {Adversarial {Machine} {Learning} at {Scale}},
	url = {http://arxiv.org/abs/1611.01236},
	doi = {10.48550/arXiv.1611.01236},
	booktitle = {{International} {Conference} {on} {Learning} {Representations} {(ICLR)} 2017},
	publisher = {arXiv},
	author = {Kurakin, Alexey and Goodfellow, Ian and Bengio, Samy},
	month = feb,
	year = {2017},
}

@inproceedings{li_shape-texture_2021,
	title = {Shape-{Texture} {Debiased} {Neural} {Network} {Training}},
	url = {http://arxiv.org/abs/2010.05981},
	language = {en},
	booktitle = {{International} {Conference} {on} {Learning} {Representations} {(ICLR)} 2021},
	publisher = {arXiv},
	author = {Li, Yingwei and Yu, Qihang and Tan, Mingxing and Mei, Jieru and Tang, Peng and Shen, Wei and Yuille, Alan and Xie, Cihang},
	month = mar,
	year = {2021},
}

@inproceedings{hendrycks_benchmarking_2019,
	title = {Benchmarking {Neural} {Network} {Robustness} to {Common} {Corruptions} and {Perturbations}},
	url = {http://arxiv.org/abs/1903.12261},
	doi = {10.48550/arXiv.1903.12261},
	booktitle = {{International} {Conference} {on} {Learning} {Representations} {(ICLR)} 2019},
	publisher = {arXiv},
	author = {Hendrycks, Dan and Dietterich, Thomas},
	month = mar,
	year = {2019},
}

@inproceedings{gatys_texture_2015,
	title = {Texture {Synthesis} {Using} {Convolutional} {Neural} {Networks}},
	url = {http://arxiv.org/abs/1505.07376},
	language = {en},
	booktitle = {{Conference} {on} {Neural} {Information} {Processing} {Systems} {(NeurIPS)} 2015},
	publisher = {arXiv},
	author = {Gatys, Leon A. and Ecker, Alexander S. and Bethge, Matthias},
	month = nov,
	year = {2015},
}

@inproceedings{bau_network_2017,
	title = {Network {Dissection}: {Quantifying} {Interpretability} of {Deep} {Visual} {Representations}},
	shorttitle = {Network {Dissection}},
	url = {http://arxiv.org/abs/1704.05796},
	doi = {10.48550/arXiv.1704.05796},
	booktitle = {{IEEE/CVF} {Conference} {on} {Computer} {Vision} {and} {Pattern} {Recognition} {(CVPR)} 2017},
	publisher = {arXiv},
	author = {Bau, David and Zhou, Bolei and Khosla, Aditya and Oliva, Aude and Torralba, Antonio},
	month = apr,
	year = {2017},
}

@inproceedings{hermann_origins_2020,
	title = {The {Origins} and {Prevalence} of {Texture} {Bias} in {Convolutional} {Neural} {Networks}},
	url = {http://arxiv.org/abs/1911.09071},
	language = {en},
	booktitle = {{Conference} {on} {Neural} {Information} {Processing} {Systems} {(NeurIPS)} 2020},
	publisher = {arXiv},
	author = {Hermann, Katherine L. and Chen, Ting and Kornblith, Simon},
	month = nov,
	year = {2020},
}

@inproceedings{liu_convnet_2022,
	title = {A {ConvNet} for the 2020s},
	url = {https://arxiv.org/abs/2201.03545v2},
	language = {en},
	booktitle = {{IEEE/CVF} {Conference} {on} {Computer} {Vision} {and} {Pattern} {Recognition} {(CVPR)} 2022},
	author = {Liu, Zhuang and Mao, Hanzi and Wu, Chao-Yuan and Feichtenhofer, Christoph and Darrell, Trevor and Xie, Saining},
	month = jan,
	year = {2022},
}

@inproceedings{szegedy_rethinking_2015,
	title = {Rethinking the {Inception} {Architecture} for {Computer} {Vision}},
	url = {http://arxiv.org/abs/1512.00567},
	doi = {10.48550/arXiv.1512.00567},
	booktitle = {{IEEE/CVF} {Conference} {on} {Computer} {Vision} {and} {Pattern} {Recognition} {(CVPR)} 2016},
	publisher = {arXiv},
	author = {Szegedy, Christian and Vanhoucke, Vincent and Ioffe, Sergey and Shlens, Jonathon and Wojna, Zbigniew},
	month = dec,
	year = {2015},
}

@inproceedings{huang_densely_2018,
	title = {Densely {Connected} {Convolutional} {Networks}},
	url = {http://arxiv.org/abs/1608.06993},
	doi = {10.48550/arXiv.1608.06993},
	booktitle = {{IEEE/CVF} {Conference} {on} {Computer} {Vision} {and} {Pattern} {Recognition} {(CVPR)} 2017},
	publisher = {arXiv},
	author = {Huang, Gao and Liu, Zhuang and van der Maaten, Laurens and Weinberger, Kilian Q.},
	month = jan,
	year = {2018},
}

@inproceedings{tan_efficientnet_2020,
	title = {{EfficientNet}: {Rethinking} {Model} {Scaling} for {Convolutional} {Neural} {Networks}},
	shorttitle = {{EfficientNet}},
	url = {http://arxiv.org/abs/1905.11946},
	doi = {10.48550/arXiv.1905.11946},
	booktitle = {{International} {Conference} {on} {Machine} {Learning} {(ICML)} 2019},
	publisher = {arXiv},
	author = {Tan, Mingxing and Le, Quoc V.},
	month = sep,
	year = {2020},
}

@inproceedings{he_deep_2015,
	title = {Deep {Residual} {Learning} for {Image} {Recognition}},
	url = {http://arxiv.org/abs/1512.03385},
	doi = {10.48550/arXiv.1512.03385},
	booktitle = {{IEEE/CVF} {Conference} {on} {Computer} {Vision} {and} {Pattern} {Recognition} {(CVPR)} 2016},
	publisher = {arXiv},
	author = {He, Kaiming and Zhang, Xiangyu and Ren, Shaoqing and Sun, Jian},
	month = dec,
	year = {2015},
}

@inproceedings{russakovsky_imagenet_2015,
	title = {{ImageNet} {Large} {Scale} {Visual} {Recognition} {Challenge}},
	url = {http://arxiv.org/abs/1409.0575},
	language = {en},
	booktitle = {{IJCV} 2015},
	publisher = {arXiv},
	author = {Russakovsky, Olga and Deng, Jia and Su, Hao and Krause, Jonathan and Satheesh, Sanjeev and Ma, Sean and Huang, Zhiheng and Karpathy, Andrej and Khosla, Aditya and Bernstein, Michael and Berg, Alexander C. and Fei-Fei, Li},
	month = jan,
	year = {2015},
}

@inproceedings{papernot_limitations_2015,
	title = {The {Limitations} of {Deep} {Learning} in {Adversarial} {Settings}},
	url = {http://arxiv.org/abs/1511.07528},
	doi = {10.48550/arXiv.1511.07528},
	booktitle = {{IEEE} {Euro} {S}\&{P} 2016},
	publisher = {arXiv},
	author = {Papernot, Nicolas and McDaniel, Patrick and Jha, Somesh and Fredrikson, Matt and Celik, Z. Berkay and Swami, Ananthram},
	month = nov,
	year = {2015},
}

@inproceedings{carlini_towards_2017,
	title = {Towards {Evaluating} the {Robustness} of {Neural} {Networks}},
	url = {http://arxiv.org/abs/1608.04644},
	doi = {10.48550/arXiv.1608.04644},
	booktitle = {{IEEE} {S}\&{P} 2017},
	publisher = {arXiv},
	author = {Carlini, Nicholas and Wagner, David},
	month = mar,
	year = {2017},
}

@inproceedings{sheatsley_space_2022,
	title = {The {Space} of {Adversarial} {Strategies}},
	copyright = {All rights reserved},
	url = {http://arxiv.org/abs/2209.04521},
	language = {en},
	booktitle = {{USENIX} {Security} 2023},
	publisher = {arXiv},
	author = {Sheatsley, Ryan and Hoak, Blaine and Pauley, Eric and McDaniel, Patrick},
	month = sep,
	year = {2022},
	note = {Number: arXiv:2209.04521
arXiv:2209.04521 [cs]},
}

@inproceedings{moosavi-dezfooli_deepfool_2016,
	title = {{DeepFool}: a simple and accurate method to fool deep neural networks},
	shorttitle = {{DeepFool}},
	url = {http://arxiv.org/abs/1511.04599},
	doi = {10.48550/arXiv.1511.04599},
	booktitle = {{IEEE/CVF} {Conference} {on} {Computer} {Vision} {and} {Pattern} {Recognition} {(CVPR)} 2016},
	publisher = {arXiv},
	author = {Moosavi-Dezfooli, Seyed-Mohsen and Fawzi, Alhussein and Frossard, Pascal},
	month = jul,
	year = {2016},
}

@inproceedings{madry_towards_2019,
	title = {Towards {Deep} {Learning} {Models} {Resistant} to {Adversarial} {Attacks}},
	url = {http://arxiv.org/abs/1706.06083},
	doi = {10.48550/arXiv.1706.06083},
	booktitle = {{International} {Conference} {on} {Learning} {Representations} {(ICLR)} 2018},
	publisher = {arXiv},
	author = {Madry, Aleksander and Makelov, Aleksandar and Schmidt, Ludwig and Tsipras, Dimitris and Vladu, Adrian},
	month = sep,
	year = {2019},
}

@inproceedings{goodfellow_explaining_2015,
	title = {Explaining and {Harnessing} {Adversarial} {Examples}},
	url = {http://arxiv.org/abs/1412.6572},
	doi = {10.48550/arXiv.1412.6572},
	booktitle = {{International} {Conference} {on} {Learning} {Representations} {(ICLR)} 2015},
	publisher = {arXiv},
	author = {Goodfellow, Ian J. and Shlens, Jonathon and Szegedy, Christian},
	month = mar,
	year = {2015},
}

@inproceedings{biggio_evasion_2013,
	title = {Evasion {Attacks} against {Machine} {Learning} at {Test} {Time}},
	url = {http://arxiv.org/abs/1708.06131},
	doi = {10.1007/978-3-642-40994-3_25},
	booktitle = {{ECML} {PKDD}},
	author = {Biggio, Battista and Corona, Igino and Maiorca, Davide and Nelson, Blaine and Srndic, Nedim and Laskov, Pavel and Giacinto, Giorgio and Roli, Fabio},
	year = {2013},
}

@inproceedings{szegedy_intriguing_2014,
	title = {Intriguing properties of neural networks},
	url = {http://arxiv.org/abs/1312.6199},
	doi = {10.48550/arXiv.1312.6199},
	booktitle = {{International} {Conference} {on} {Learning} {Representations} {(ICLR)} 2014},
	publisher = {arXiv},
	author = {Szegedy, Christian and Zaremba, Wojciech and Sutskever, Ilya and Bruna, Joan and Erhan, Dumitru and Goodfellow, Ian and Fergus, Rob},
	month = feb,
	year = {2014},
}

@inproceedings{hoak_explorations_2024,
	title = {Explorations in {Texture} {Learning}},
	url = {http://arxiv.org/abs/2403.09543},
	doi = {10.48550/arXiv.2403.09543},
	booktitle = {{International} {Conference} {on} {Learning} {Representations} {(ICLR)} 2024, {Tiny} {Papers} {Track}},
	publisher = {arXiv},
	author = {Hoak, Blaine and McDaniel, Patrick},
	month = mar,
	year = {2024},
}

@inproceedings{hendrycks_natural_2021,
	title = {Natural {Adversarial} {Examples}},
	url = {http://arxiv.org/abs/1907.07174},
	doi = {10.48550/arXiv.1907.07174},
	booktitle = {{IEEE/CVF} {Conference} {on} {Computer} {Vision} {and} {Pattern} {Recognition} {(CVPR)} 2021},
	publisher = {arXiv},
	author = {Hendrycks, Dan and Zhao, Kevin and Basart, Steven and Steinhardt, Jacob and Song, Dawn},
	month = mar,
	year = {2021},
}

@inproceedings{geirhos_generalisation_2020,
	title = {Generalisation in humans and deep neural networks},
	url = {http://arxiv.org/abs/1808.08750},
	doi = {10.48550/arXiv.1808.08750},
	booktitle = {{Conference} {on} {Neural} {Information} {Processing} {Systems} {(NeurIPS)} 2018},
	publisher = {arXiv},
	author = {Geirhos, Robert and Temme, Carlos R. Medina and Rauber, Jonas and Schütt, Heiko H. and Bethge, Matthias and Wichmann, Felix A.},
	month = oct,
	year = {2020},
}

@inproceedings{brendel_approximating_2019,
	title = {Approximating {CNNs} with {Bag}-of-local-{Features} models works surprisingly well on {ImageNet}},
	url = {http://arxiv.org/abs/1904.00760},
	doi = {10.48550/arXiv.1904.00760},
	booktitle = {{International} {Conference} {on} {Learning} {Representations} {(ICLR)} 2019},
	publisher = {arXiv},
	author = {Brendel, Wieland and Bethge, Matthias},
	month = mar,
	year = {2019},
}

@inproceedings{geirhos_imagenet-trained_2019,
	title = {{ImageNet}-trained {CNNs} are biased towards texture; increasing shape bias improves accuracy and robustness},
	url = {http://arxiv.org/abs/1811.12231},
	language = {en},
	booktitle = {{International} {Conference} {on} {Learning} {Representations} {(ICLR)}},
	author = {Geirhos, Robert and Rubisch, Patricia and Michaelis, Claudio and Bethge, Matthias and Wichmann, Felix A. and Brendel, Wieland},
	month = jan,
	year = {2019},
}

@article{rafegas_color_2018,
	series = {Color: cone opponency and beyond},
	title = {Color encoding in biologically-inspired convolutional neural networks},
	volume = {151},
	issn = {0042-6989},
	url = {https://www.sciencedirect.com/science/article/pii/S0042698918300592},
	doi = {10.1016/j.visres.2018.03.010},
	journal = {Vision Research},
	author = {Rafegas, Ivet and Vanrell, Maria},
	month = oct,
	year = {2018},
	pages = {7--17},
}
% \bibliography{references}
\appendix\section{Appendix}\label{appendix}
% \subsection{Experimental Details}\label{appendix_details}
% The pretrained ResNet50 used in our experiments was obtained from torchvision~\cite{marcel_torchvision_2010} with the default model weights. The model was trained on ImageNet~\cite{russakovsky_imagenet_2015}. The model was evaluated on the DTD dataset \cite{cimpoi_describing_2014} using the following data preprocessing steps: (1) resize the image to 256$\times$256, (2) center crop the image to 224$\times$224, (3) normalize the image using the mean and standard deviation of the ImageNet training dataset. All experiments were run across 12 NVIDIA A100 GPUs. Complete code to replicate experiments can be found at \url{anonymized_for_submission}.

\subsection{Human Evaluation Details}\label{appendix:human_eval}
Here we provide the exact instructions given to the human evaluators for the validation of the TID in \autoref{sec:human_eval}.
\begin{tcolorbox}[colback=gray!10, colframe=gray!80, title=User Study Instructions, breakable]

Thank you for participating in this study! Please follow the steps below to complete the evaluation.

\textbf{Prerequisites}
\begin{itemize}
    \item You will need Python 3 installed on your machine along with the following packages. If you need to install these packages, you can do so with the following command:
    
    \texttt{pip install pandas pillow matplotlib argparse}
    \item Ensure you have internet access for downloading the package and uploading the results.
\end{itemize}

\textbf{Steps}
\begin{enumerate}

    \item \textbf{Download the Package}

    Download the provided tarball package of your choosing, the \texttt{eval\_packages.py} script, and the \texttt{README\_humaneval.md} file. These files contain the necessary instructions, images, and the script you will run for this study.

    Once you have downloaded everything, move all 3 items to a directory of your choosing.

    \item \textbf{Run the Script}

    Open a terminal in the directory where you placed the files and run the Python script with \texttt{\{package\_num\}} being the package number (shown in the tarball name) you would like to evaluate.

    \texttt{python3 eval\_packages.py {package\_num}}

    The script will display 100 images, one at a time in a pop-up window along with four words in the terminal.

    \item \textbf{Input Your Responses}

    For each image, you will be shown four texture words. Your task is to input the number corresponding to the texture that you believe is most prominent in the image. Note that you do not have to click out of the current texture image; inputting your answer will advance to the next image.

    \begin{itemize}
        \item If you are unfamiliar with any texture word, feel free to look up examples. \href{https://www.robots.ox.ac.uk/~vgg/data/dtd/view.html?categ=banded}{This website} has many (but not all) of the textures from this study and is a great resource. To see other textures, either click next on the website or change the last word in the link to the desired texture.
        \item If you think multiple textures are present, you can input multiple numbers separated by spaces (e.g., \texttt{1 3}).
        \item If you feel like none are present in the image, choose the one that seems the most probable given the other information in the image.
        \item If you want to quit at any time, press \texttt{q} to exit. The script will save your progress, so you can continue from where you left off later.
    \end{itemize}

    \item \textbf{Complete the Study}

    Once you have completed the evaluation for all 100 images, a completion message will show, and the script will save your results in a CSV file.

    \item \textbf{Upload Your Results}

    Please upload the generated CSV file to the provided Google Drive link.

\end{enumerate}

\end{tcolorbox}

\subsection{Additional TID examples}\label{appendix:tid_examples}

\begin{figure}[h]
    \centering
    \includegraphics[width=\linewidth]{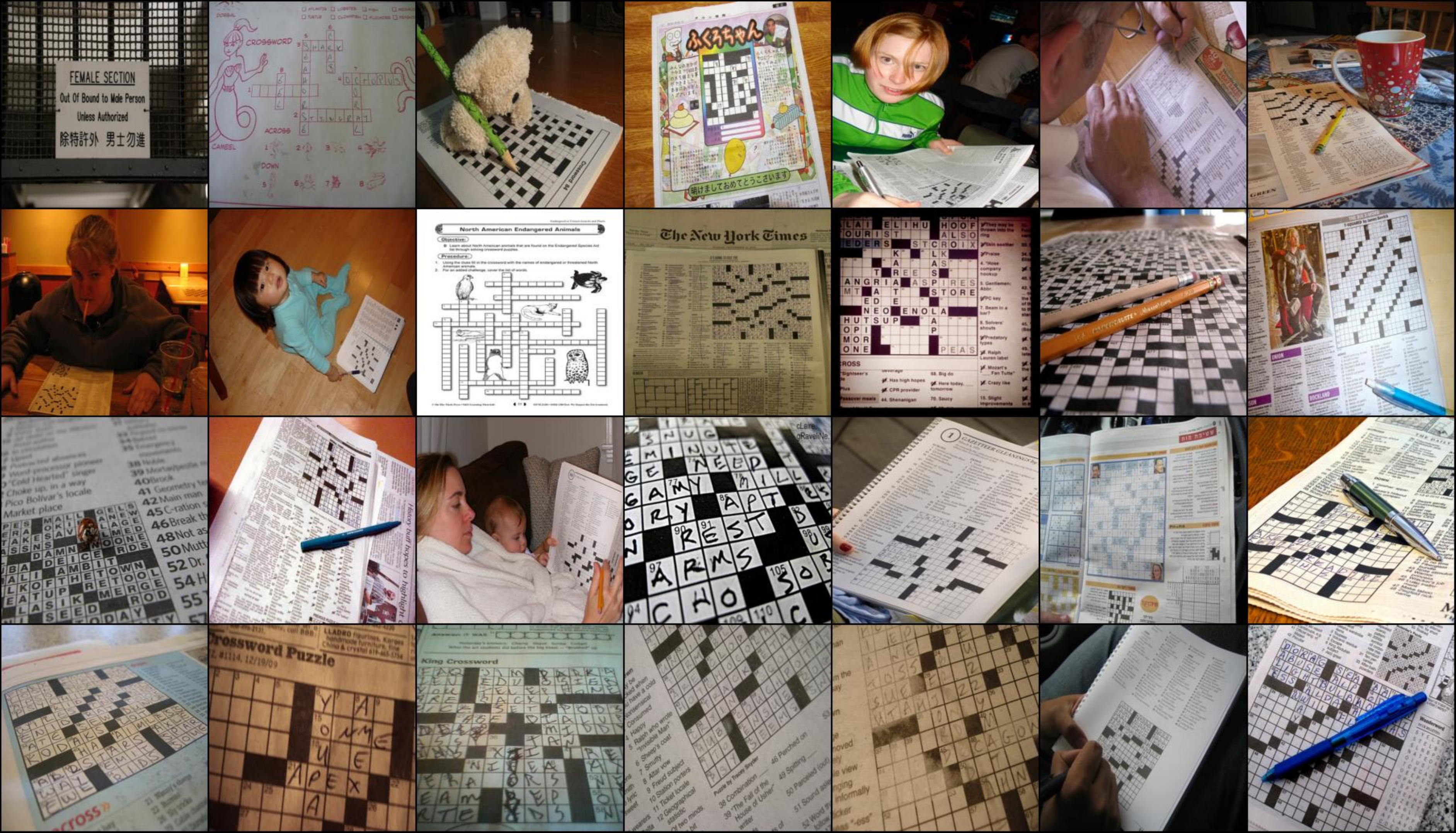}
    \caption{Images identified by TID as having a checkered texture.}
    \label{fig:tid_checkered}
\end{figure}

\begin{figure}[h]
    \centering
    \includegraphics[width=\linewidth]{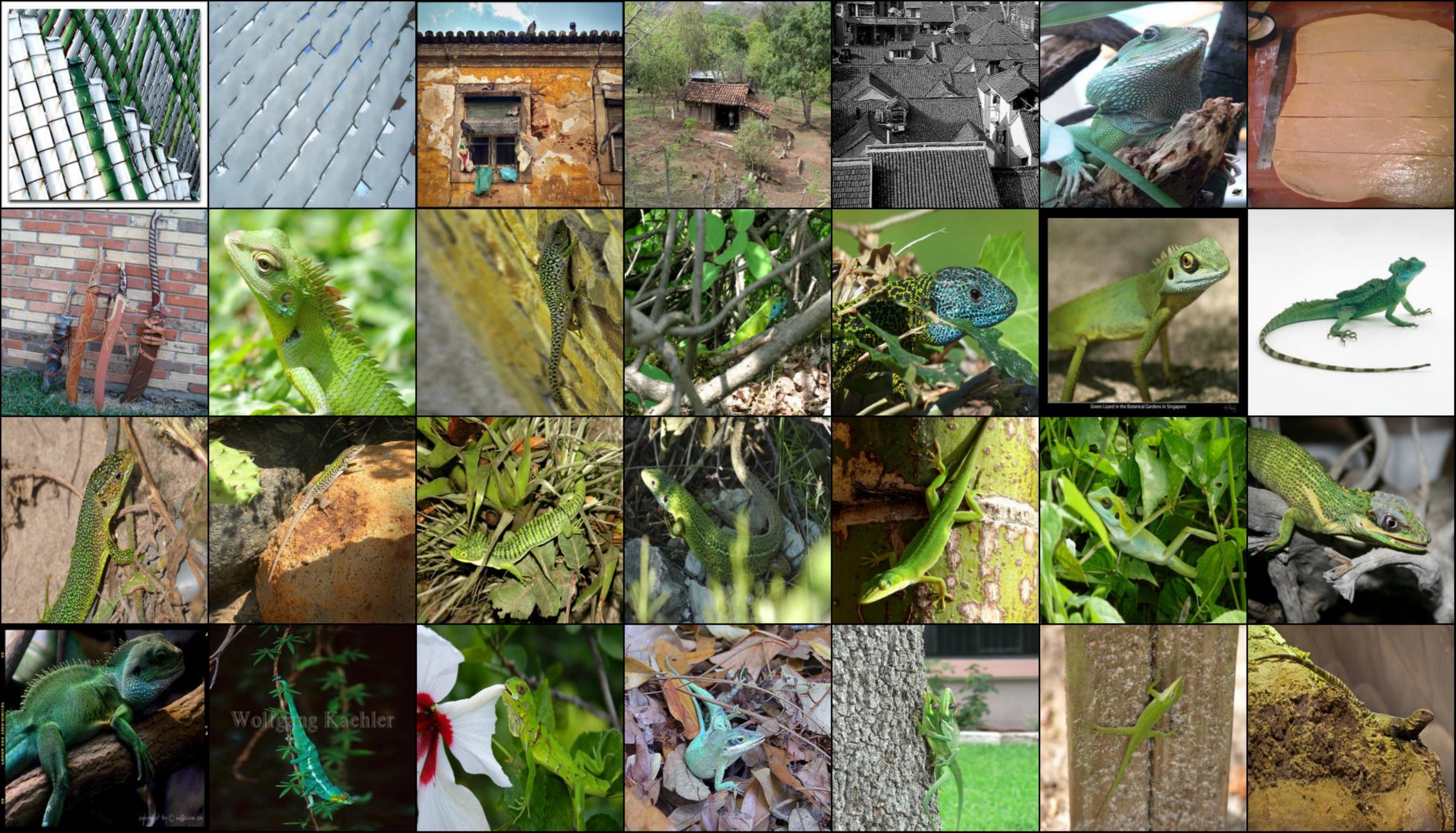}
    \caption{Images identified by TID as having a scaly texture.}
    \label{fig:tid_scaly}
\end{figure}

\begin{figure}[h]
    \centering
    \includegraphics[width=\linewidth]{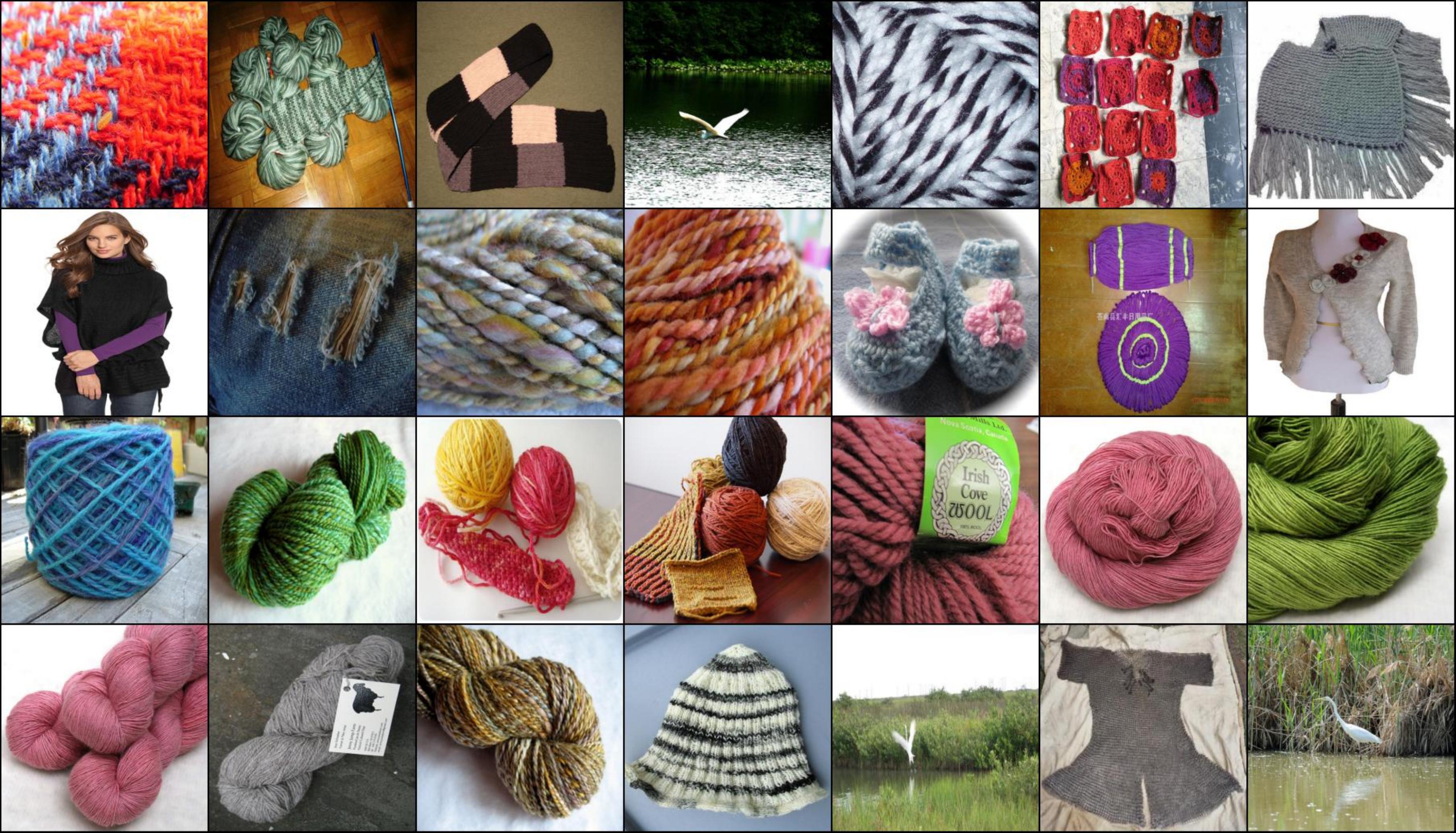}
    \caption{Images identified by TID as having a fibrous texture.}
    \label{fig:tid_fibrous}
\end{figure}

\begin{figure}[h]
    \centering
    \includegraphics[width=\linewidth]{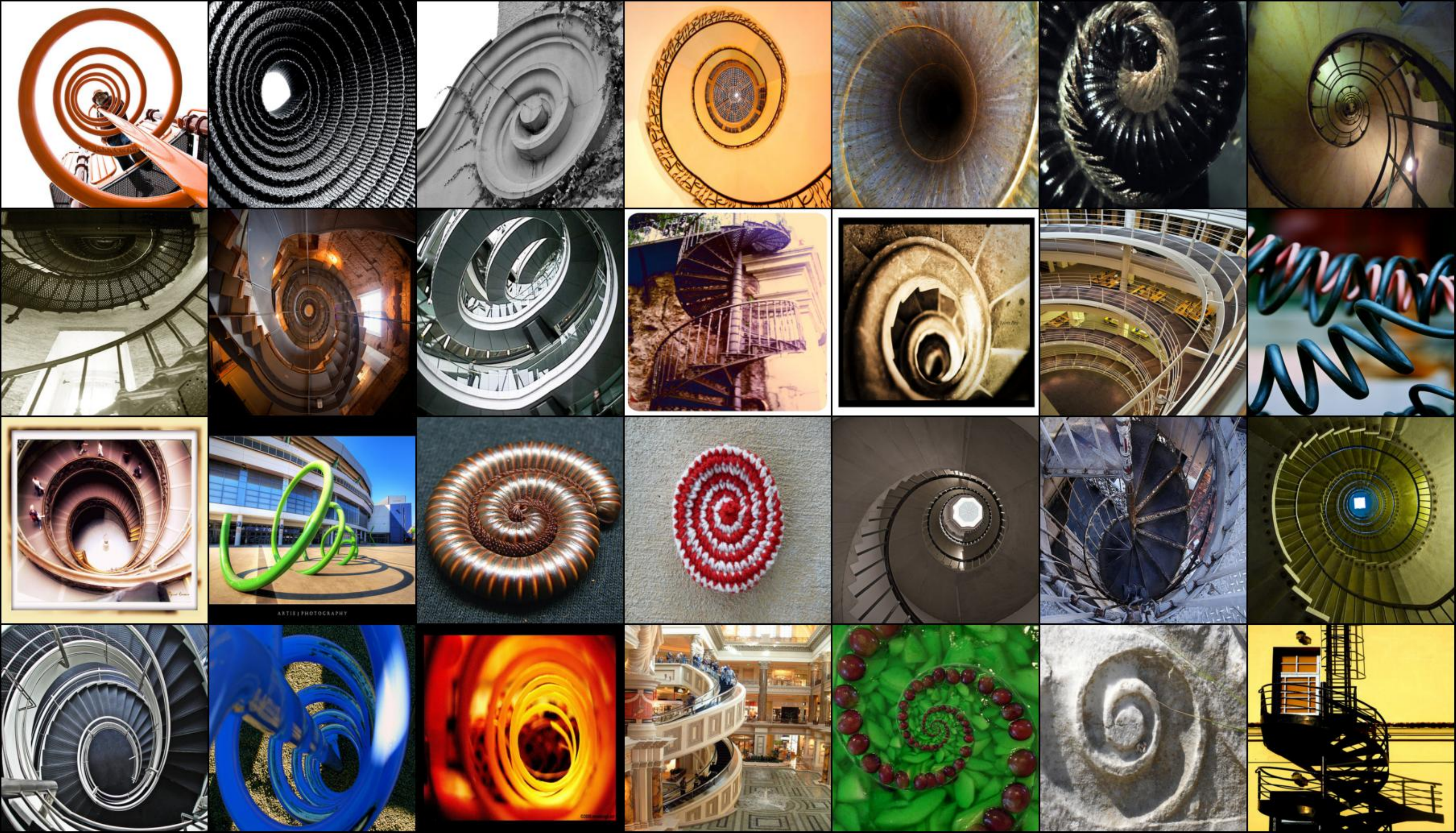}
    \caption{Images identified by TID as having a spiraled texture.}
    \label{fig:tid_spiraled}
\end{figure}

\begin{figure}[h]
    \centering
    \includegraphics[width=\linewidth]{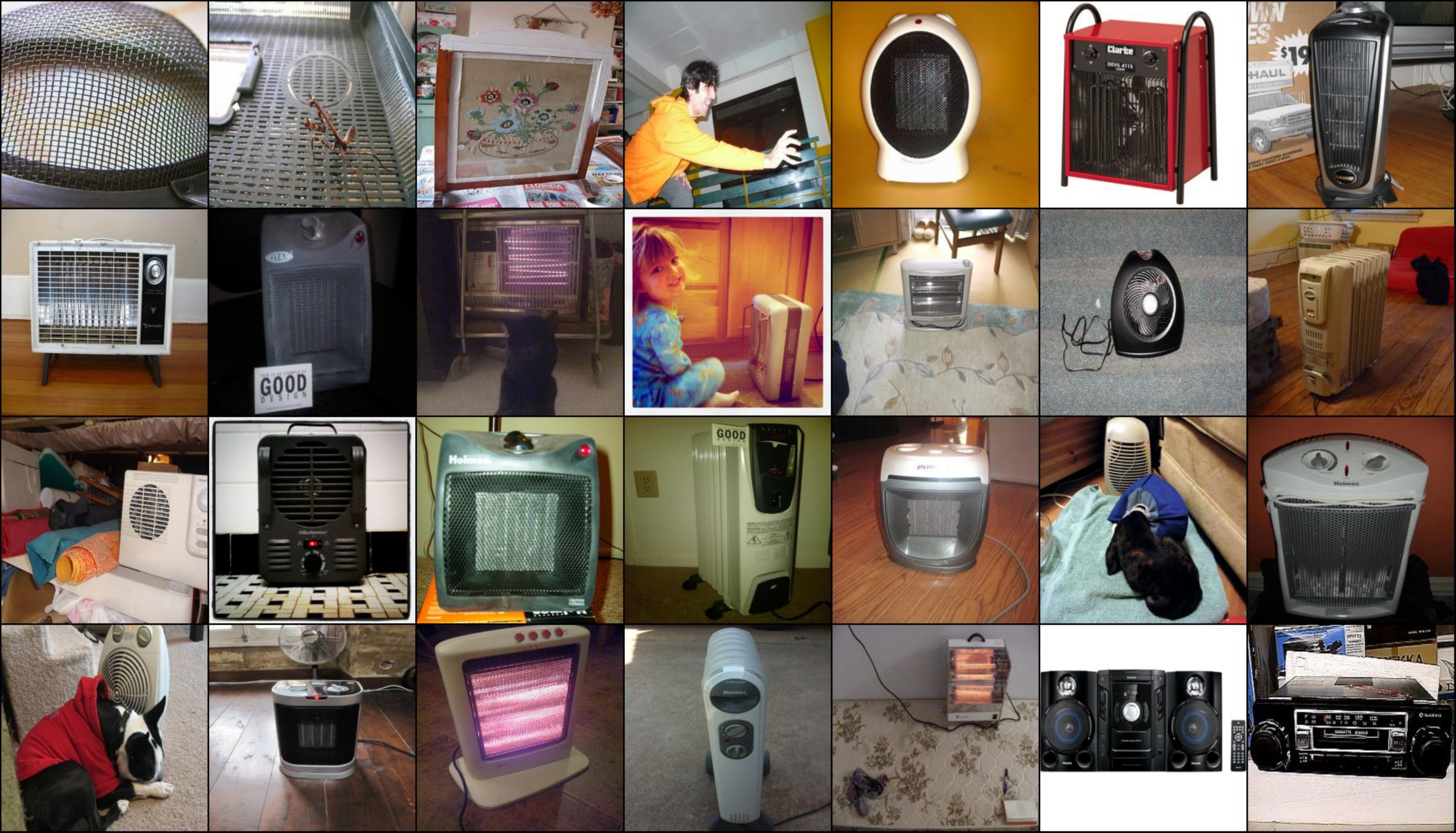}
    \caption{Images identified by TID as having a perforated texture.}
    \label{fig:tid_perf}
\end{figure}

\autoref{fig:tid_checkered},  \autoref{fig:tid_scaly}, \autoref{fig:tid_fibrous},  \autoref{fig:tid_spiraled}, and \autoref{fig:tid_perf} show examples of ImageNet validation images identified by the TID of ResNet50 as having various textures.

\subsection{ImageNet-A accuracy}

\autoref{tab:imneta_acc} displays the accuracy of each model on the ImageNet-A dataset.

\begin{table}[ht]
    \centering
    \caption{Accuracy on ImageNet-A}
    \label{tab:imneta_acc}
    \begin{tabular}{l|r}
\toprule
Model & Accuracy (\%) \\
\midrule
convnext-base & 17.04 \\
densenet121 & 0.52 \\
densenet169 & 0.96 \\
efficientnet-b0 & 2.71 \\
inception-v3 & 3.72 \\
resnet18 & 0.29 \\
resnet50 & 0.00 \\
resnet152 & 9.68 \\
\bottomrule
\end{tabular}
\end{table}

\subsection{Model confidence on texture data}\label{appendix:confidence_hist}
The following figures display the confidence histograms of different models on the Prompted Textures Dataset. Results on ResNet50 can be found in the main body of the paper in \autoref{fig:confidence_hist}.

\begin{figure}[ht]
  \centering
  \includegraphics[width=\linewidth]{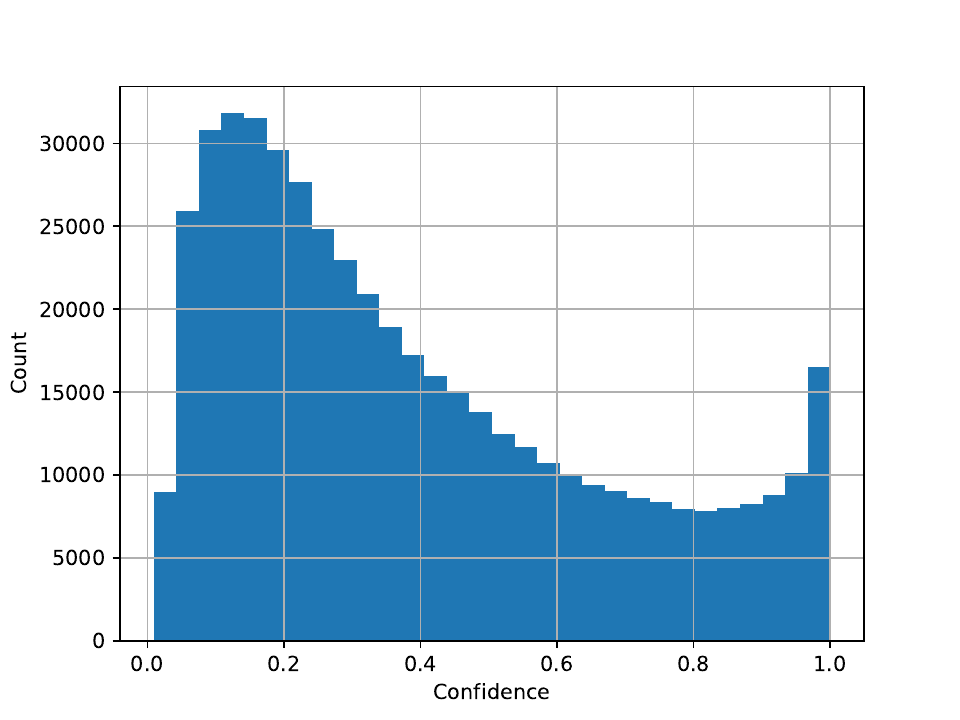}
  \caption{Confidence histogram of texture images on ResNet18.}
  \label{fig:confidence_hist_resnet18}
\end{figure}

\begin{figure}[ht]
  \centering
  \includegraphics[width=\linewidth]{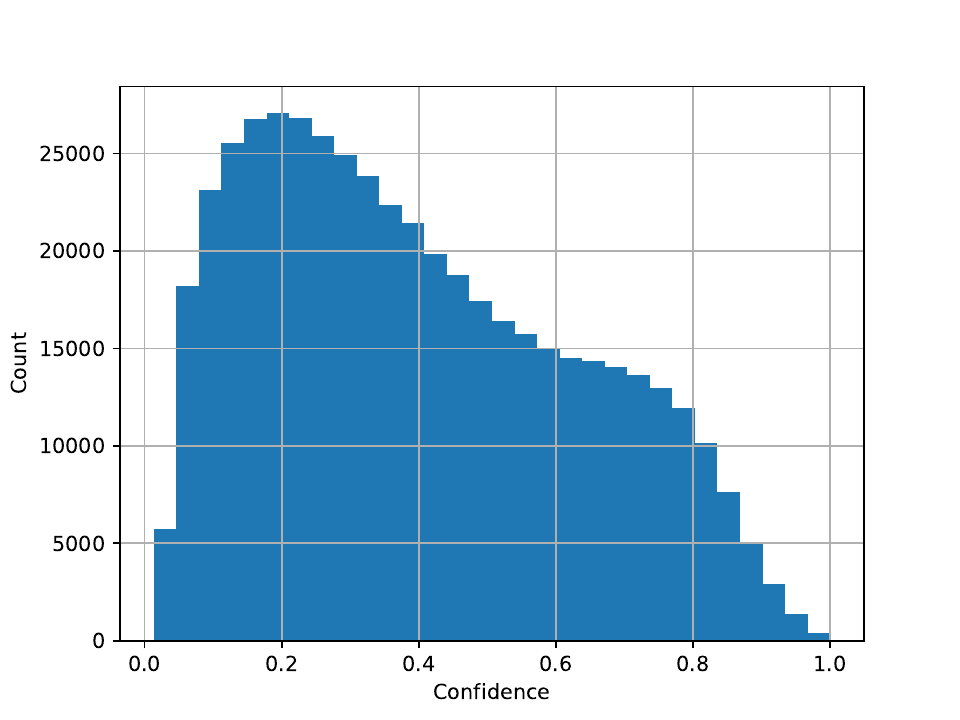}
  \caption{Confidence histogram of texture images on ResNet152.}
  \label{fig:confidence_hist_resnet152}
\end{figure}

\begin{figure}[ht]
  \centering
  \includegraphics[width=\linewidth]{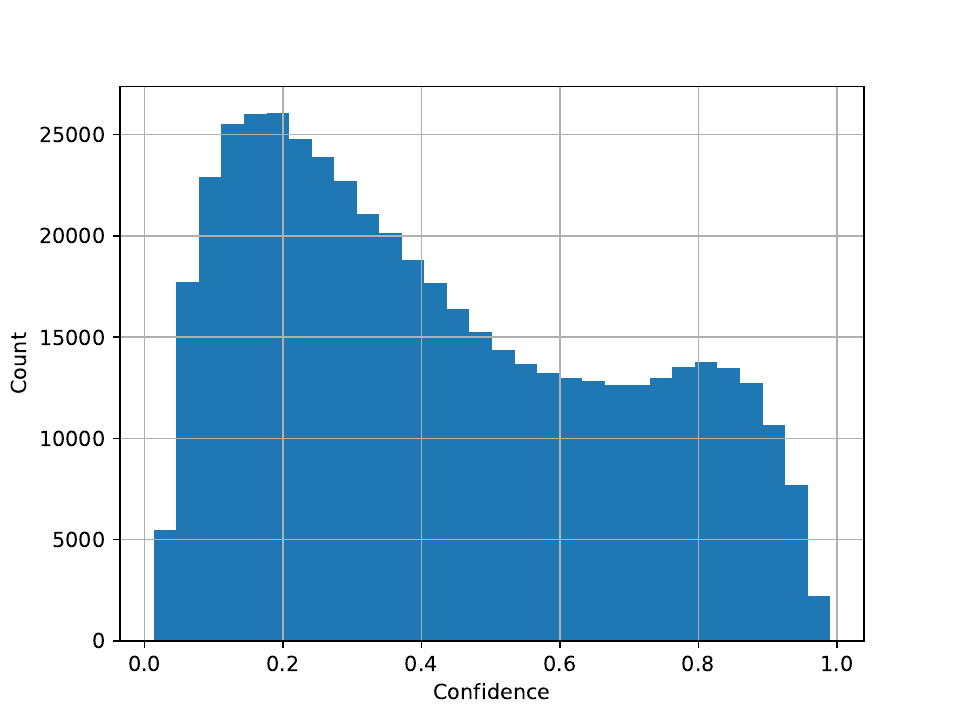}
  \caption{Confidence histogram of texture images on ConvNeXT.}
  \label{fig:confidence_hist_convnext}
\end{figure}

\begin{figure}[ht]
  \centering
  \includegraphics[width=\linewidth]{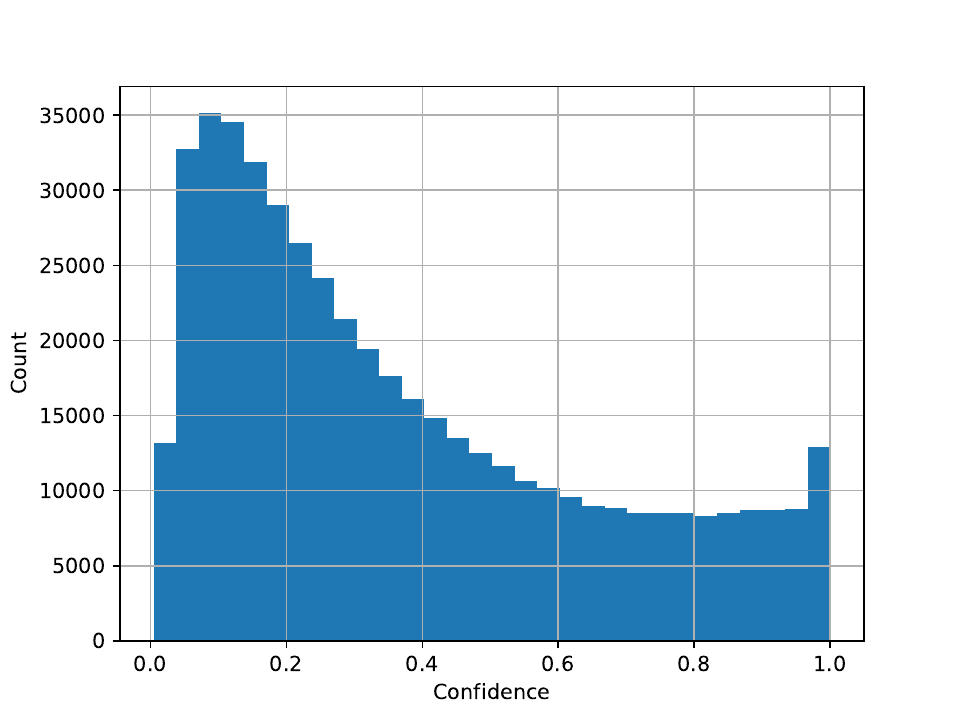}
  \caption{Confidence histogram of texture images on Inception-v3.}
  \label{fig:confidence_hist_inceptionv3}
\end{figure}

\begin{figure}[ht]
  \centering
  \includegraphics[width=\linewidth]{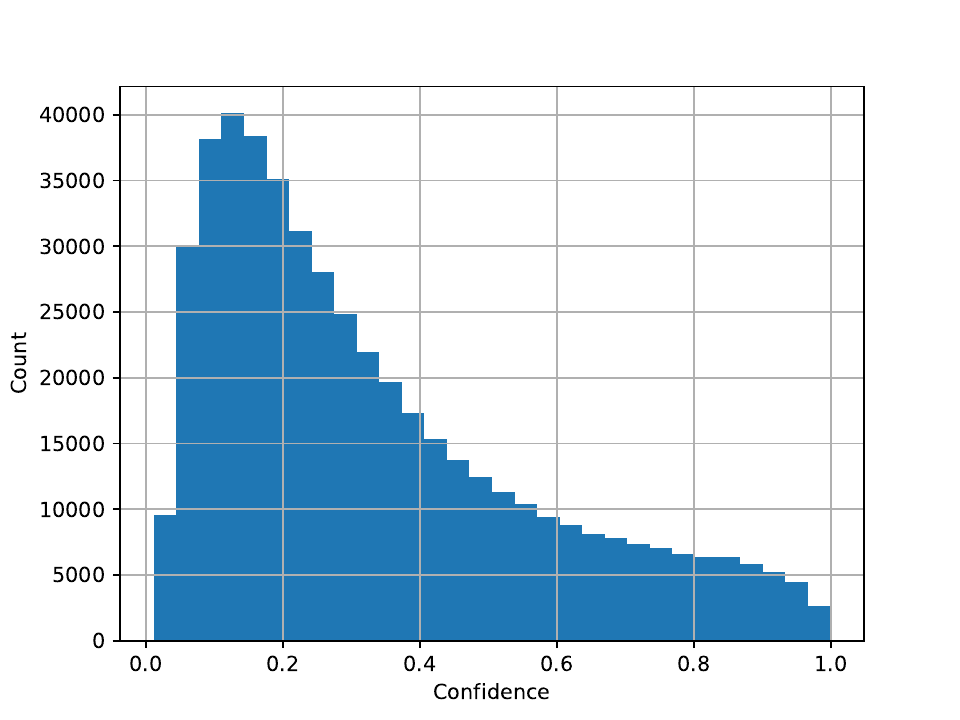}
  \caption{Confidence histogram of texture images on EfficientNet-B0.}
  \label{fig:confidence_hist_effnet}
\end{figure}

\begin{figure}[ht]
  \centering
  \includegraphics[width=\linewidth]{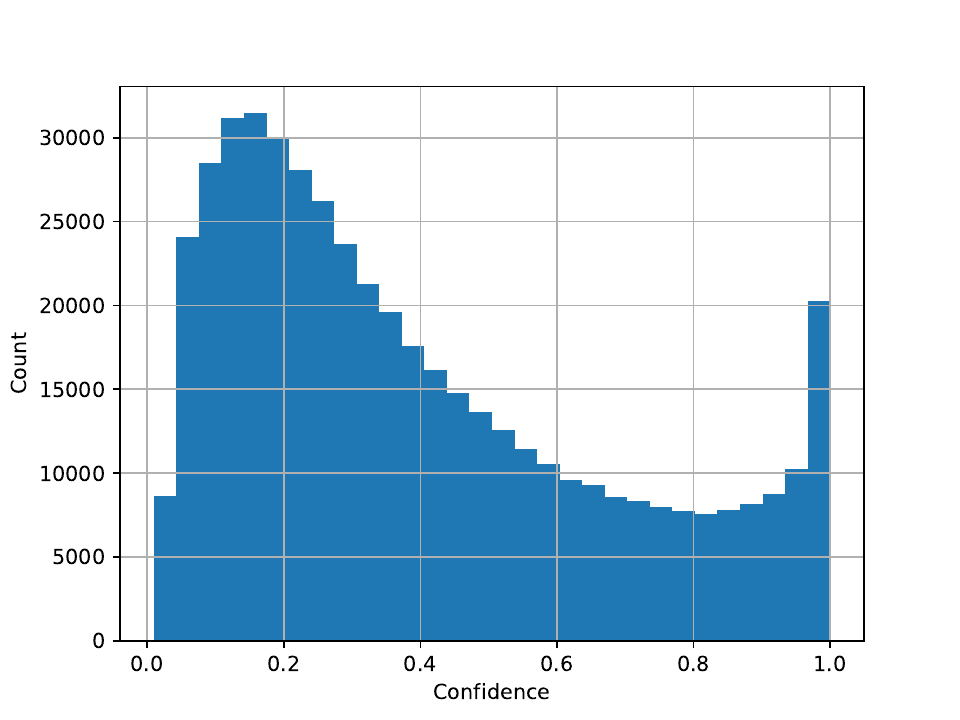}
  \caption{Confidence histogram of texture images on DenseNet121.}
  \label{fig:confidence_hist_densenet121}
\end{figure}

\begin{figure}[ht]
  \centering
  \includegraphics[width=\linewidth]{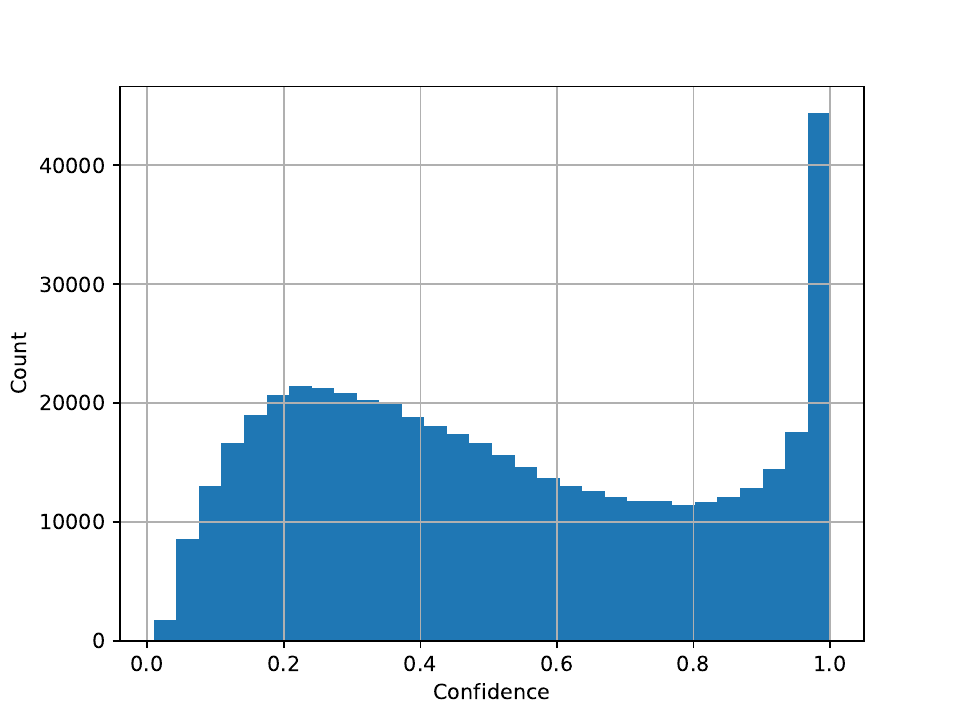}
  \caption{Confidence histogram of texture images on DenseNet169.}
  \label{fig:confidence_hist_densenet169}
\end{figure}

\newpage\subsection{Top texture object associations}\label{appendix:pairs_bar}
The following figures display the top 50 \tav{} texture-object pairs on various models. Results on ResNet50 can be found in the main body of the paper in \autoref{fig:text_obj_bar_pairs}.

\begin{figure}[ht]
  \centering
  \includegraphics[width=\linewidth]{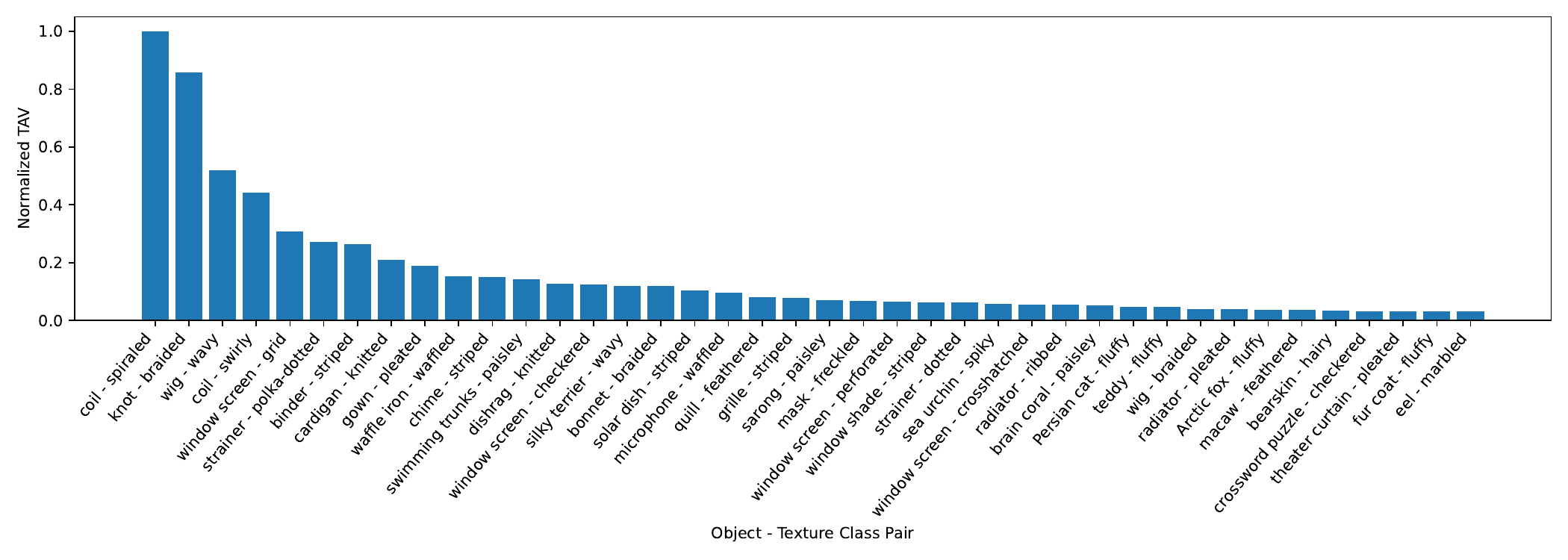}
  \caption{Top 50 strongest \tav{} pairs on ResNet18.}
  \label{fig:pairs_bar_resnet18}
\end{figure}

\begin{figure}[ht]
  \centering
  \includegraphics[width=\linewidth]{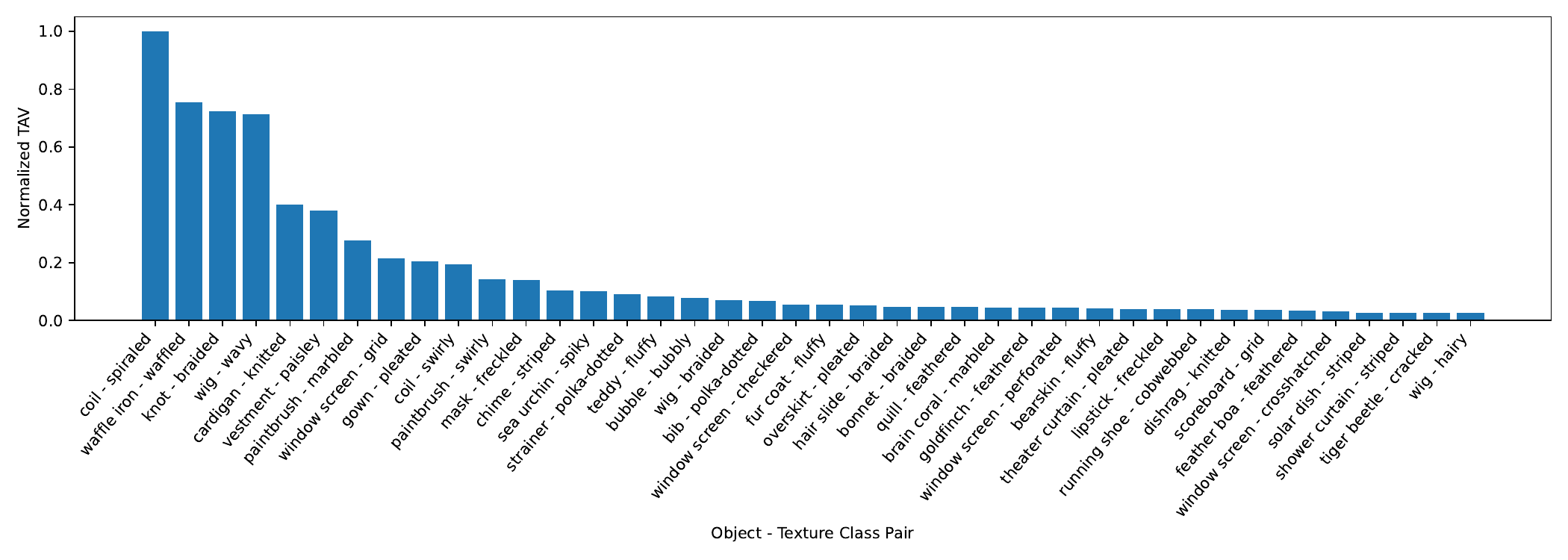}
  \caption{Top 50 strongest \tav{} pairs on ResNet152.}
  \label{fig:pairs_bar_resnet152}
\end{figure}

\begin{figure}[ht]
  \centering
  \includegraphics[width=\linewidth]{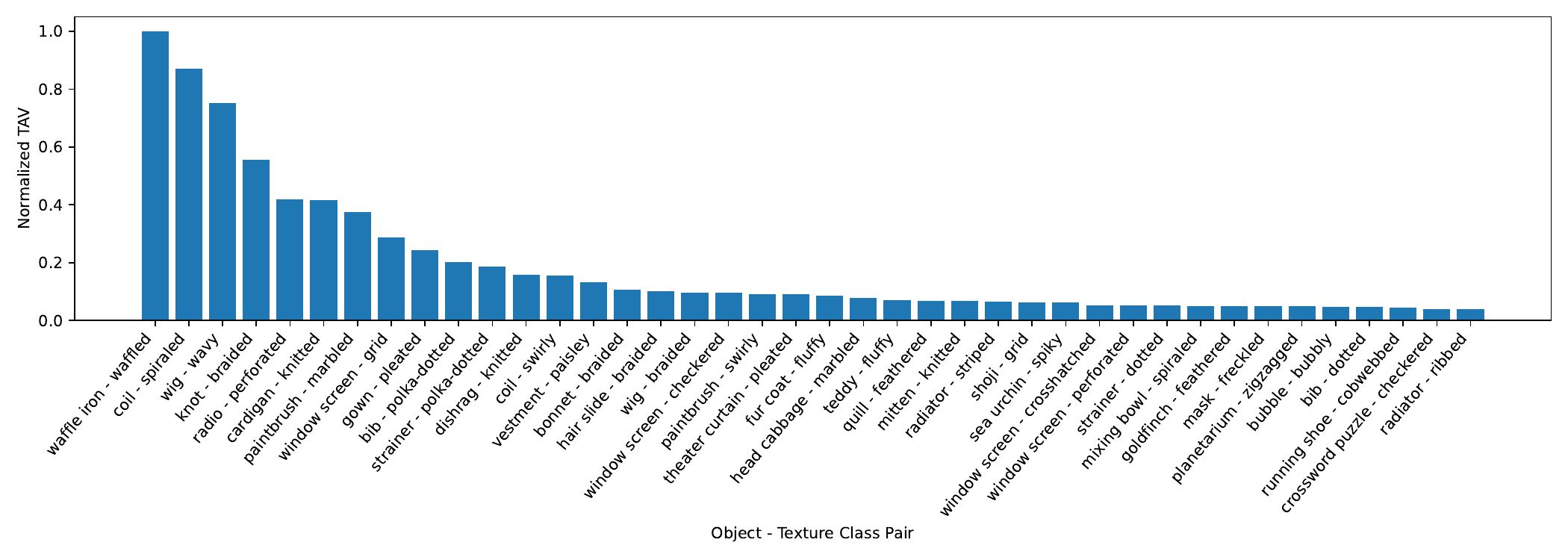}
  \caption{Top 50 strongest \tav{} pairs on ConvNeXT.}
  \label{fig:pairs_bar_convnext}
\end{figure}

\begin{figure}[ht]
  \centering
  \includegraphics[width=\linewidth]{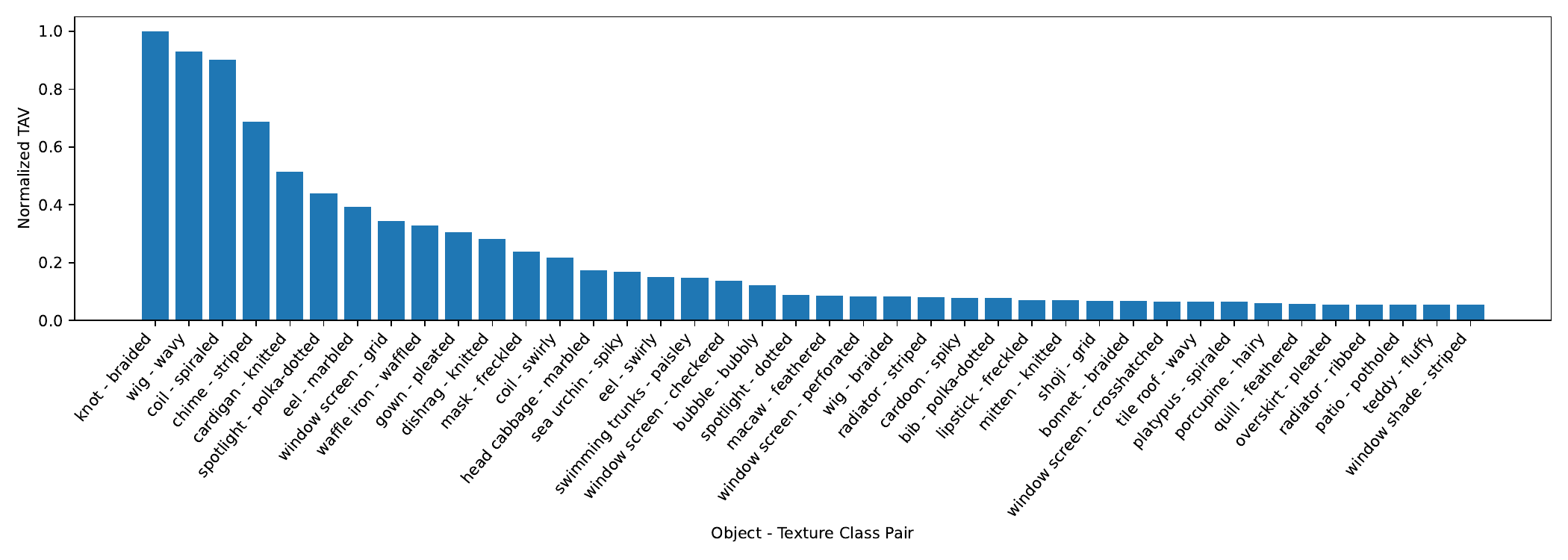}
  \caption{Top 50 strongest \tav{} pairs on Inception-v3.}
  \label{fig:pairs_bar_inceptionv3}
\end{figure}

\begin{figure}[ht]
  \centering
  \includegraphics[width=\linewidth]{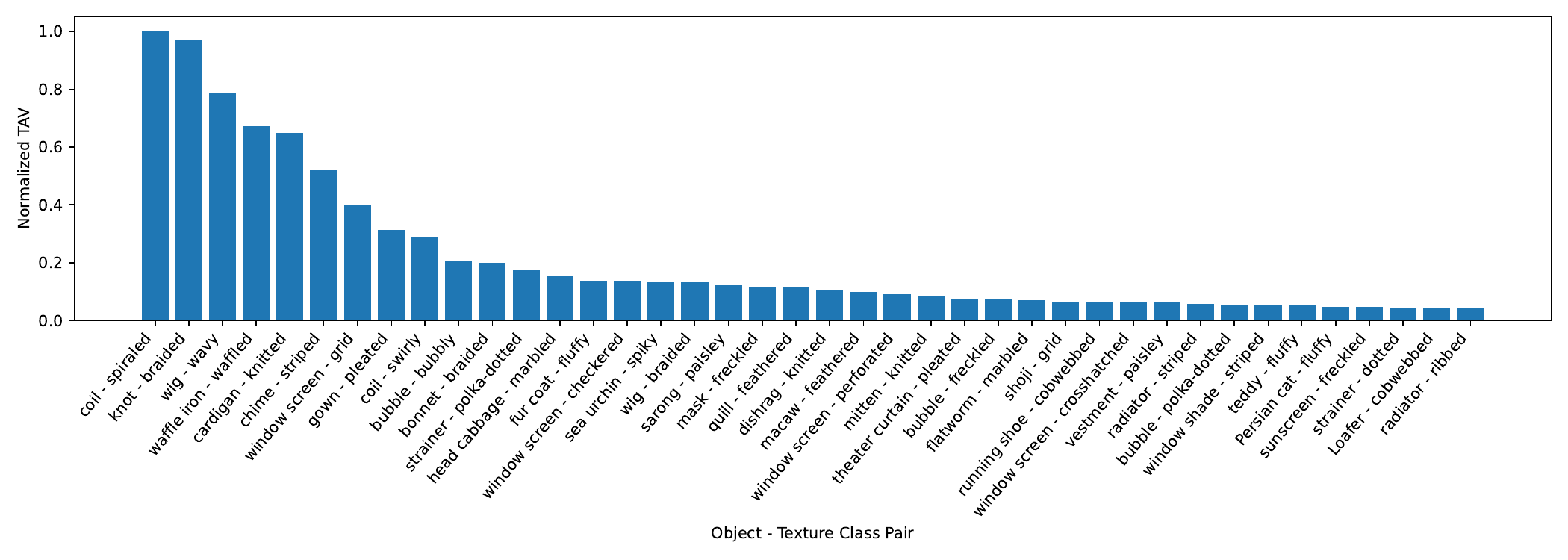}
  \caption{Top 50 strongest \tav{} pairs on EfficientNet-B0.}
  \label{fig:pairs_bar_effnet}
\end{figure}

\begin{figure}[ht]
  \centering
  \includegraphics[width=\linewidth]{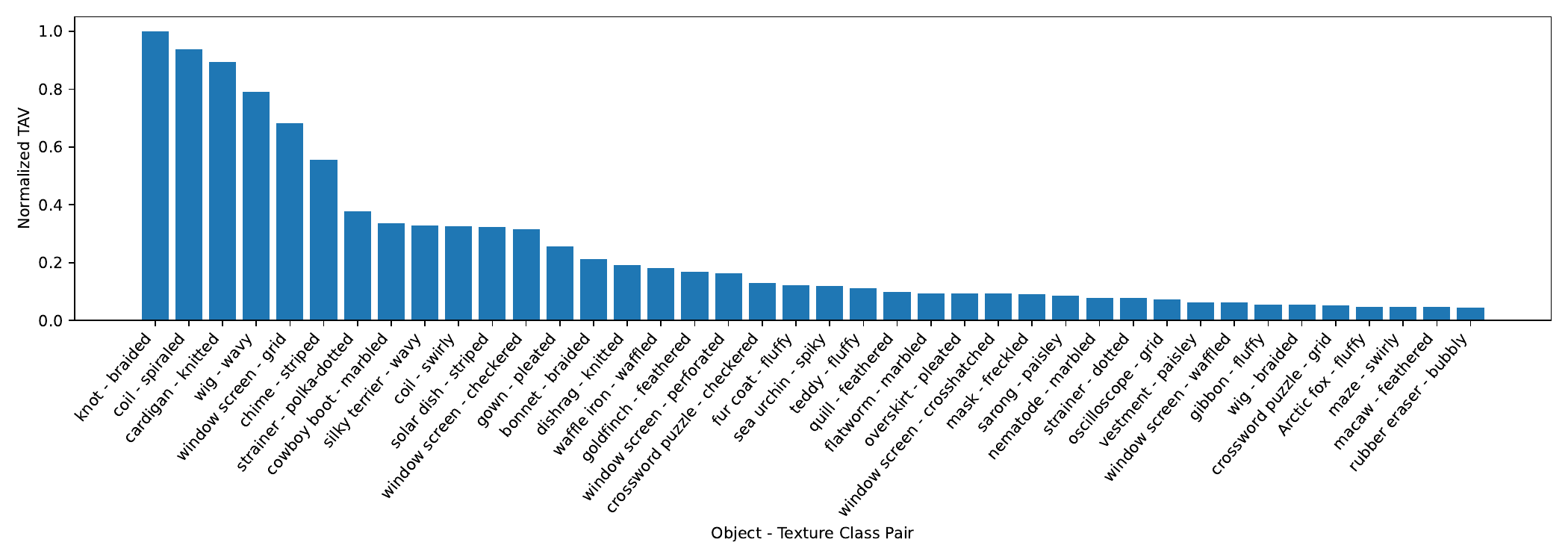}
  \caption{Top 50 strongest \tav{} pairs on DenseNet121.}
  \label{fig:pairs_bar_densenet121}
\end{figure}

\begin{figure}[ht]
  \centering
  \includegraphics[width=\linewidth]{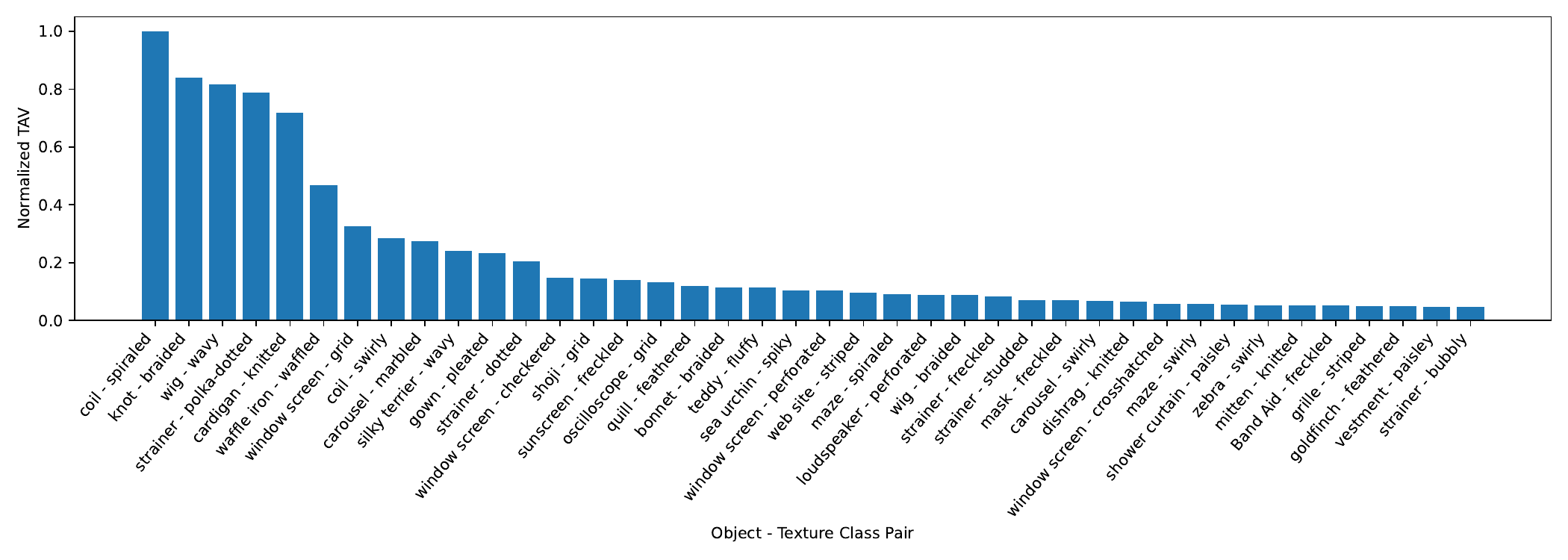}
  \caption{Top 50 strongest \tav{} pairs on DenseNet169.}
  \label{fig:pairs_bar_densenet169}
\end{figure}

\newpage\subsection{Model accuracy on different textures}\label{appendix:count_labels_accuracy_scatter}
The following figures display the average accuracy of various models on different texture groupings present in each label class by how many samples are in each group (normalized by the number of samples in each object label group) on the ImageNet validation set. Results on ResNet50 can be found in the main body of the paper in \autoref{fig:texture_object_count_labels_accuracy}. 

\begin{figure}[ht]
  \centering
  \includegraphics[width=\linewidth]{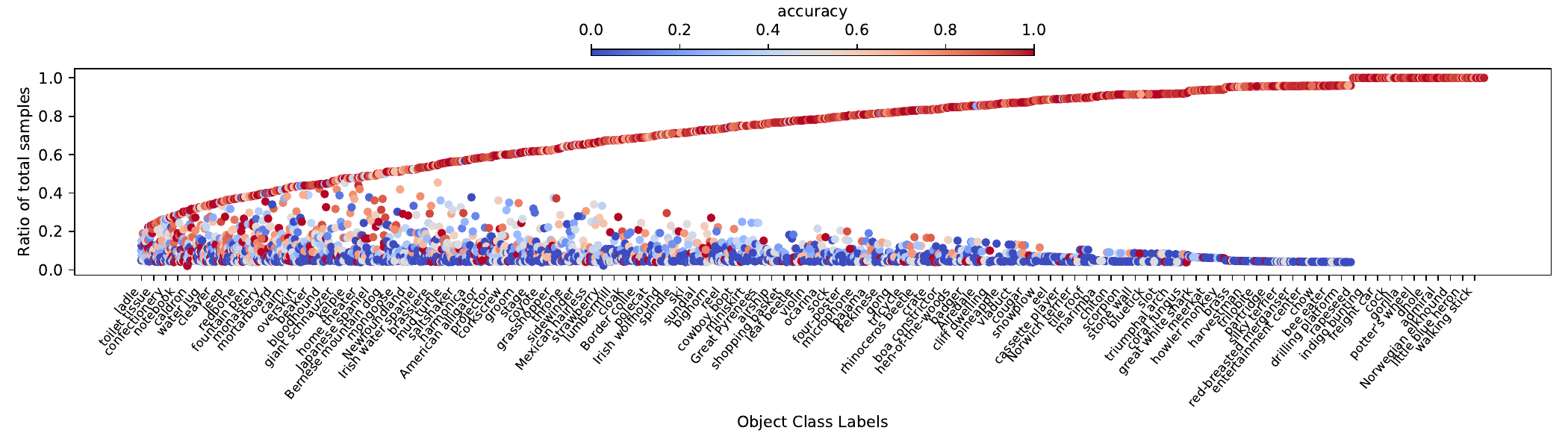}
  \caption{ResNet18.}
  \label{fig:count_labels_accuracy_scatter_resnet18}
\end{figure}

\begin{figure}[ht]
  \centering
  \includegraphics[width=\linewidth]{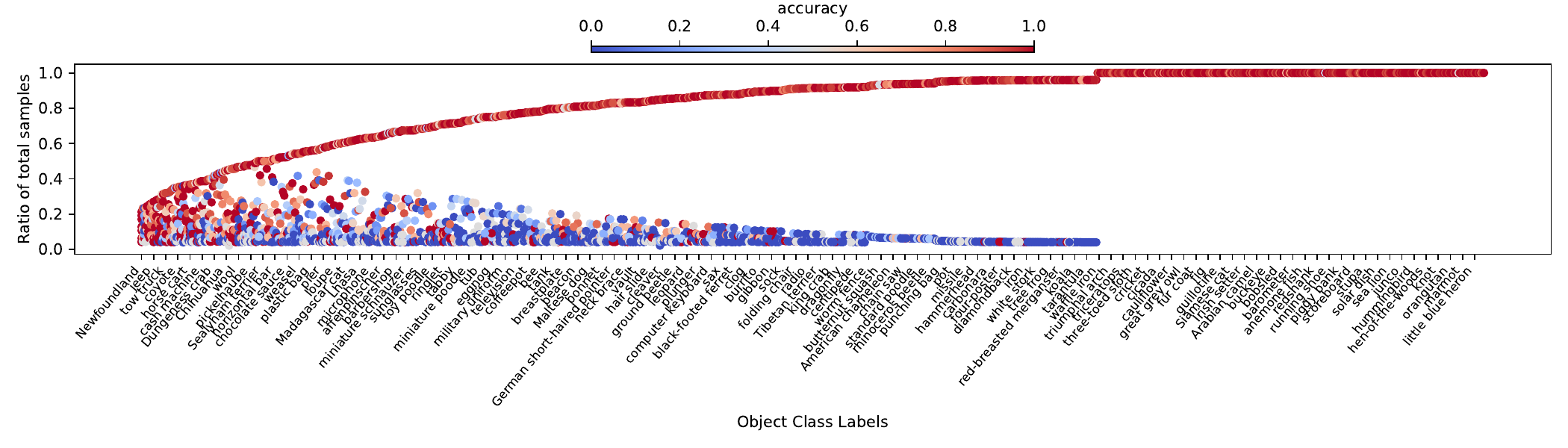}
  \caption{ResNet152.}
  \label{fig:count_labels_accuracy_scatter_resnet152}
\end{figure}

\begin{figure}[ht]
  \centering
  \includegraphics[width=\linewidth]{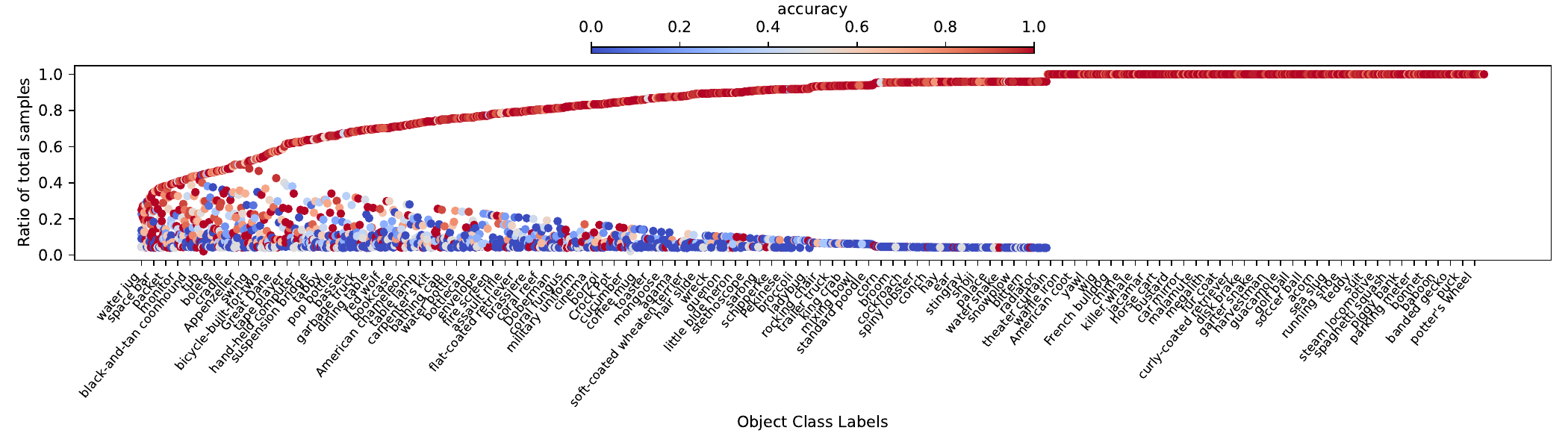}
  \caption{ConvNeXT.}
  \label{fig:count_labels_accuracy_scatter_convnext}
\end{figure}

\begin{figure}[ht]
  \centering
  \includegraphics[width=\linewidth]{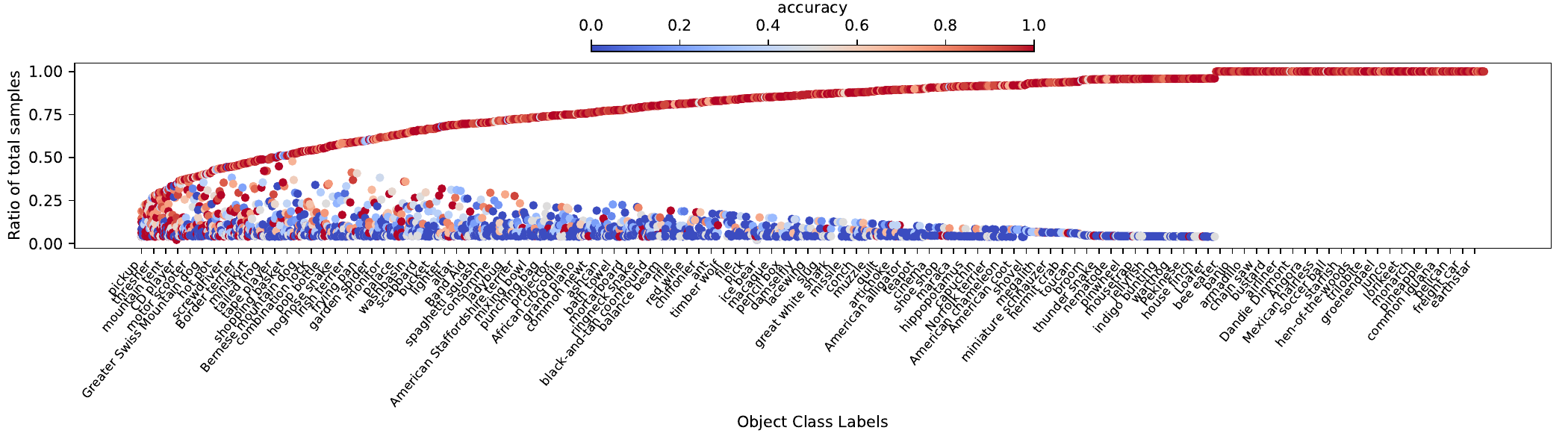}
  \caption{Inception-v3.}
  \label{fig:count_labels_accuracy_scatter_inceptionv3}
\end{figure}

\begin{figure}[ht]
  \centering
  \includegraphics[width=\linewidth]{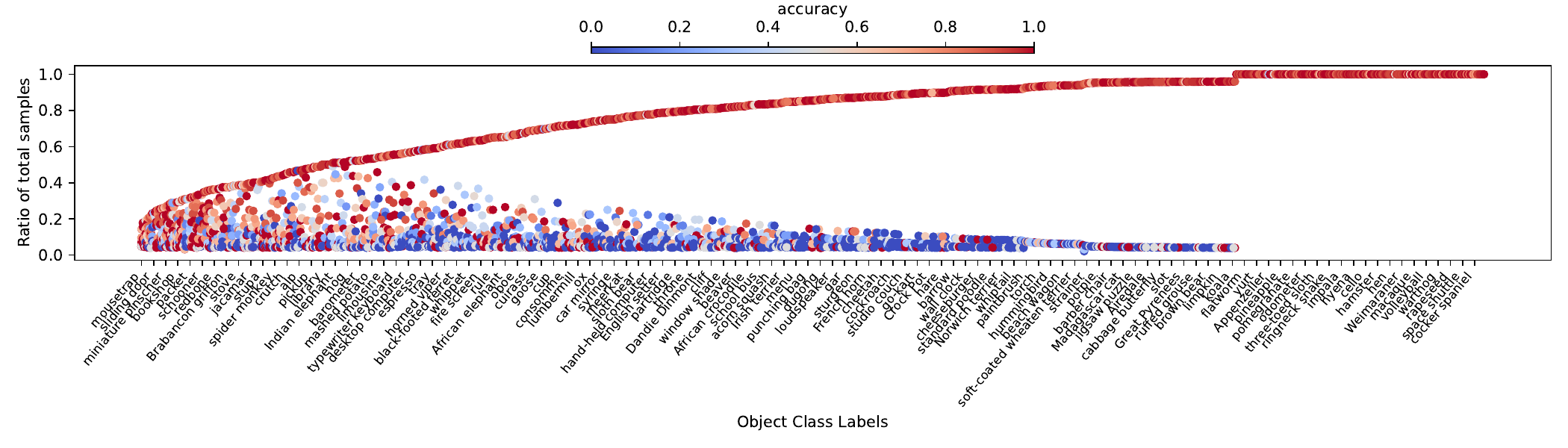}
  \caption{EfficientNet-B0.}
  \label{fig:count_labels_accuracy_scatter_effnet}
\end{figure}

\begin{figure}[ht]
  \centering
  \includegraphics[width=\linewidth]{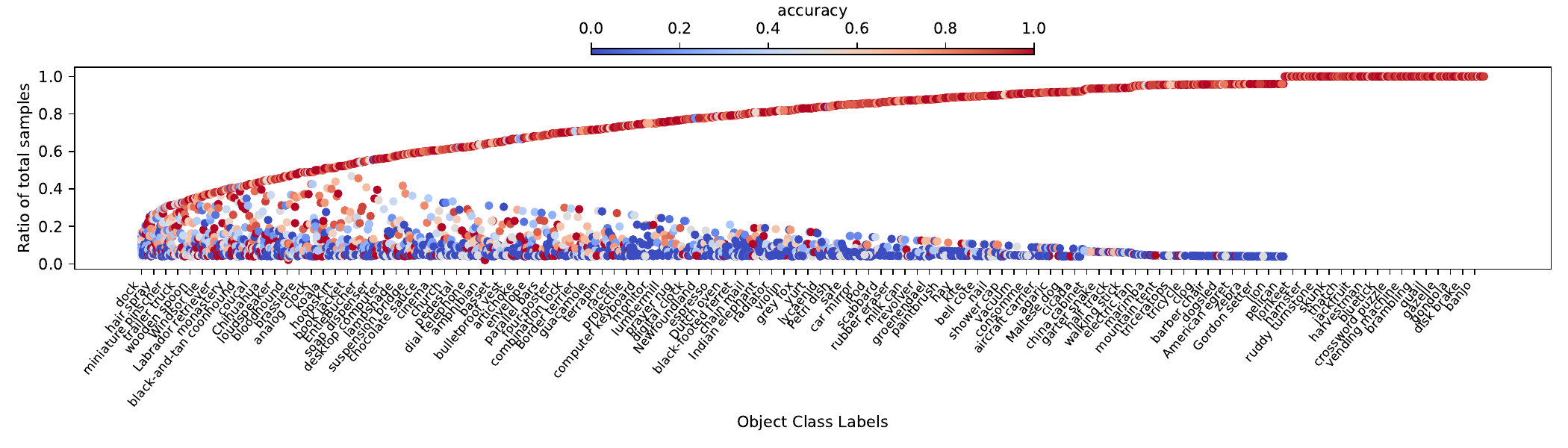}
  \caption{DenseNet121.}
  \label{fig:count_labels_accuracy_scatter_densenet121}
\end{figure}

\begin{figure}[ht]
  \centering
  \includegraphics[width=\linewidth]{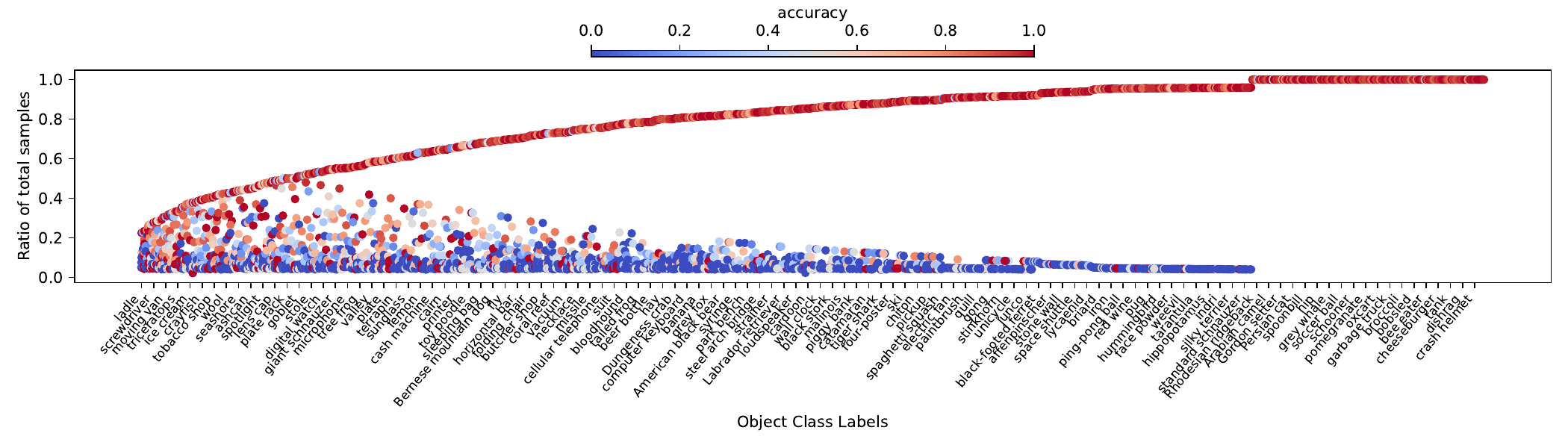}
  \caption{DenseNet169.}
  \label{fig:count_labels_accuracy_scatter_densenet169}
\end{figure}

\newpage\subsection{Model confidence on different textures}\label{appendix:count_preds_confidence_scatter}
The following figures display the average confidence of various models on different texture groupings present in each object prediction class by how many samples are in each group (normalized by the number of samples in each object prediction group) on the ImageNet validation set. Results on ResNet50 can be found in the main body of the paper in \autoref{fig:texture_object_count_preds_conf}.

\begin{figure}[ht]
  \centering
  \includegraphics[width=\linewidth]{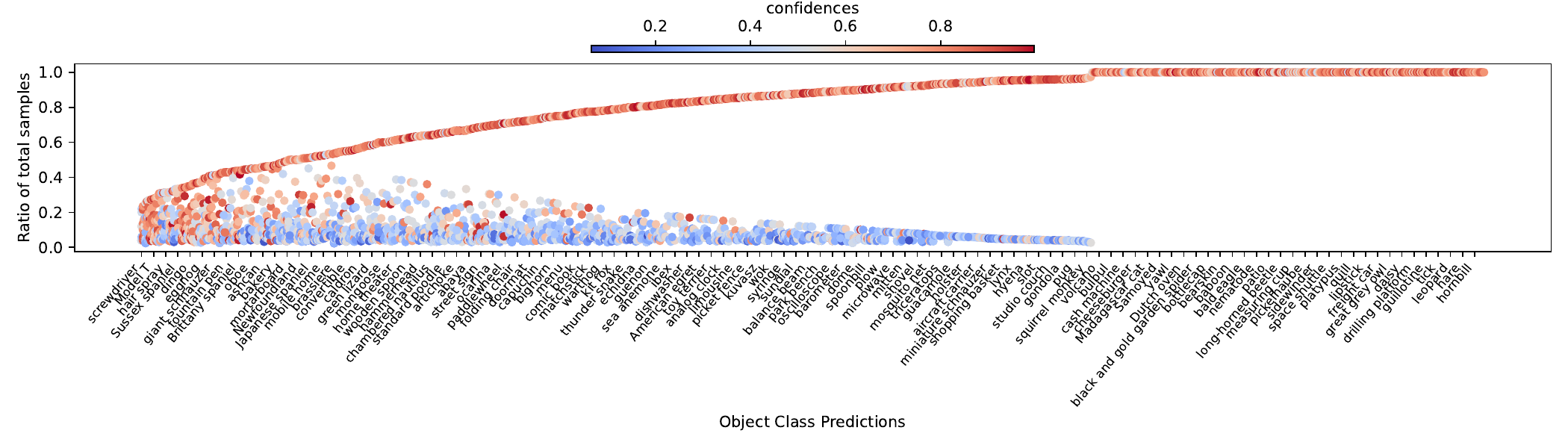}
  \caption{ResNet18.}
  \label{fig:count_preds_conf_scatter_resnet18}
\end{figure}

\begin{figure}[ht]
  \centering
  \includegraphics[width=\linewidth]{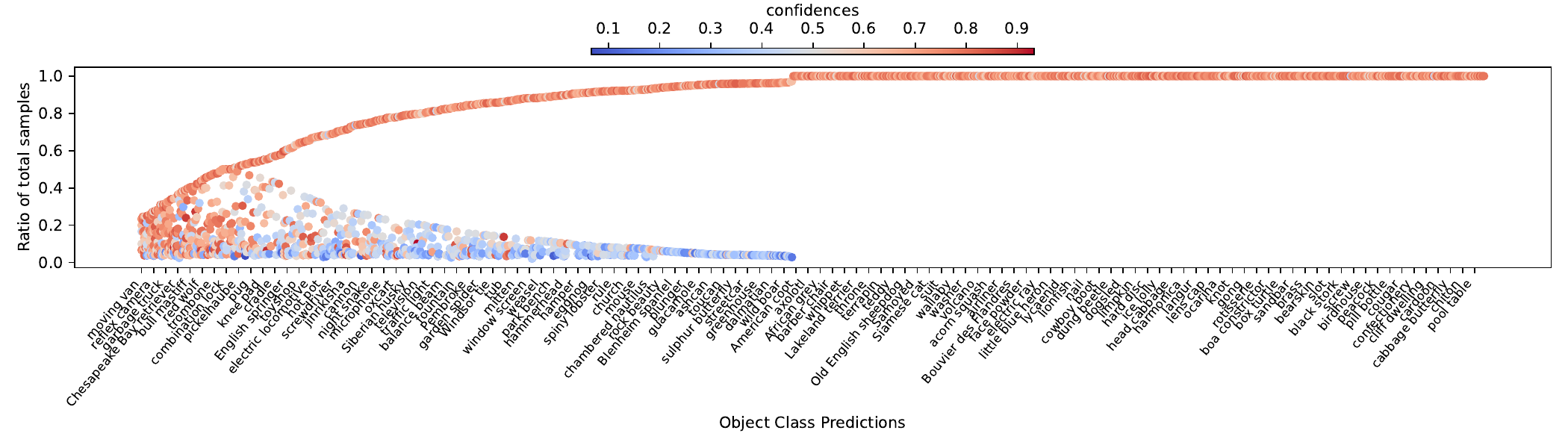}
  \caption{ResNet152.}
  \label{fig:count_preds_conf_scatter_resnet152}
\end{figure}

\begin{figure}[ht]
  \centering
  \includegraphics[width=\linewidth]{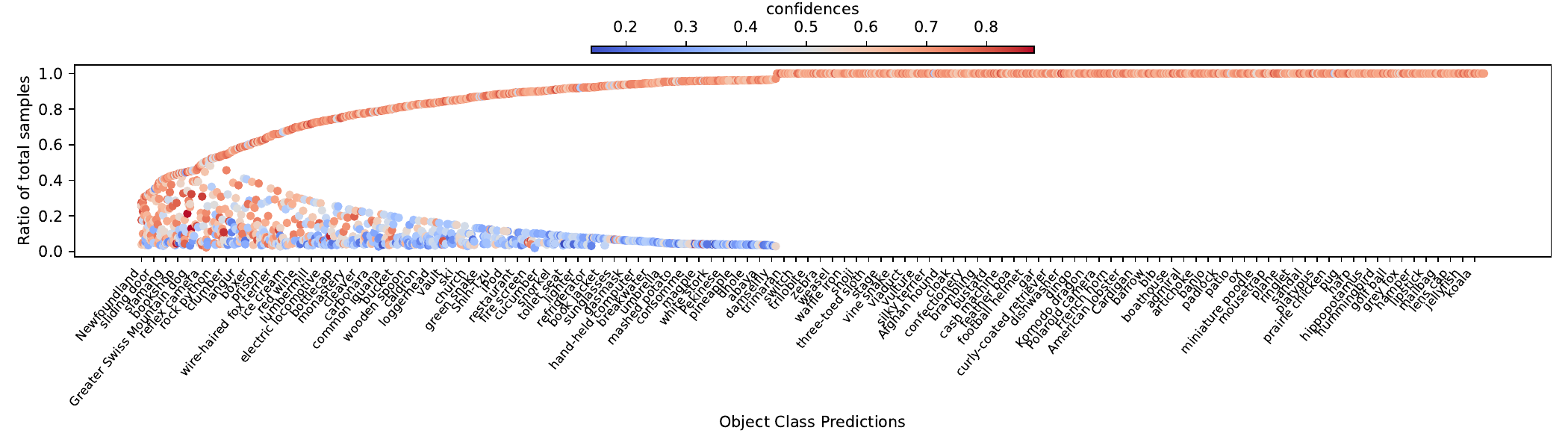}
  \caption{ConvNeXT.}
  \label{fig:count_preds_conf_scatter_convnext}
\end{figure}

\begin{figure}[ht]
  \centering
  \includegraphics[width=\linewidth]{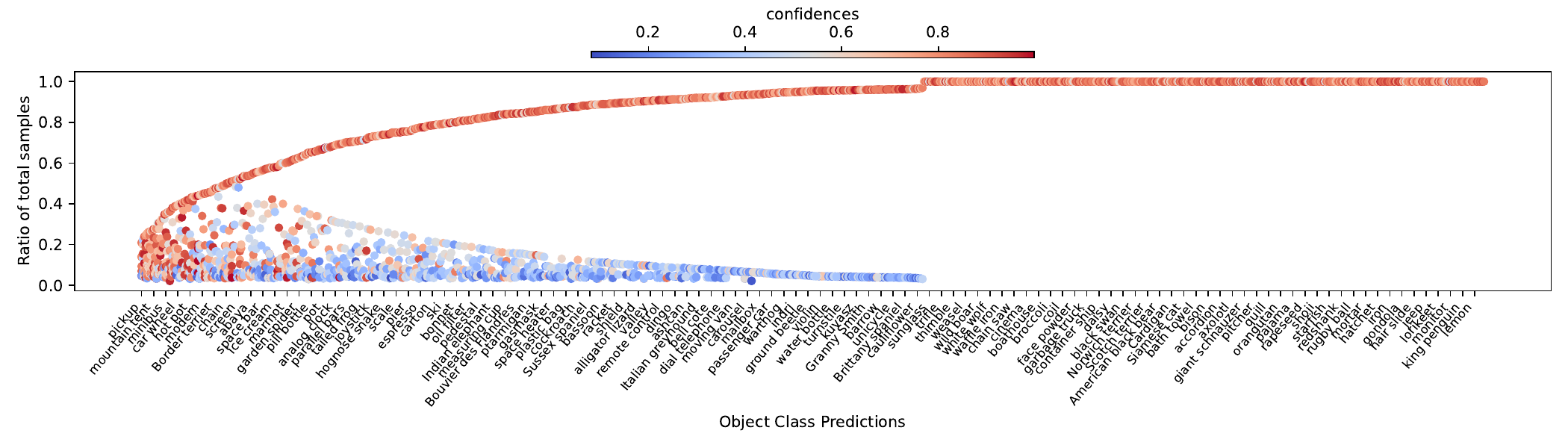}
  \caption{Inception-v3.}
  \label{fig:count_preds_conf_scatter_inceptionv3}
\end{figure}

\begin{figure}[ht]
  \centering
  \includegraphics[width=\linewidth]{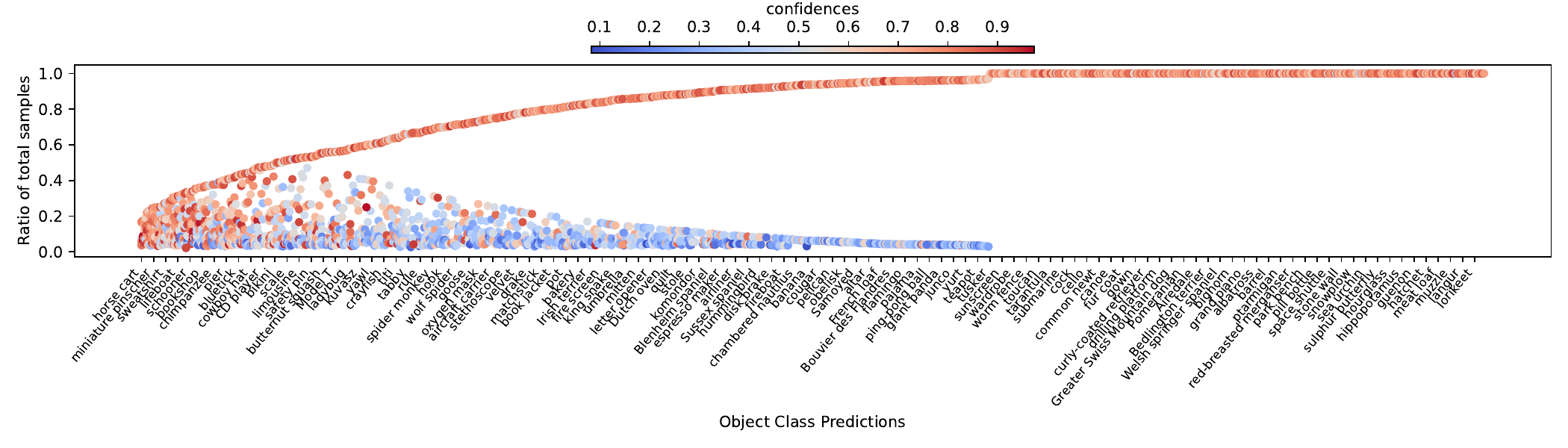}
  \caption{EfficientNet-B0.}
  \label{fig:count_preds_conf_scatter_effnet}
\end{figure}

\begin{figure}[ht]
  \centering
  \includegraphics[width=\linewidth]{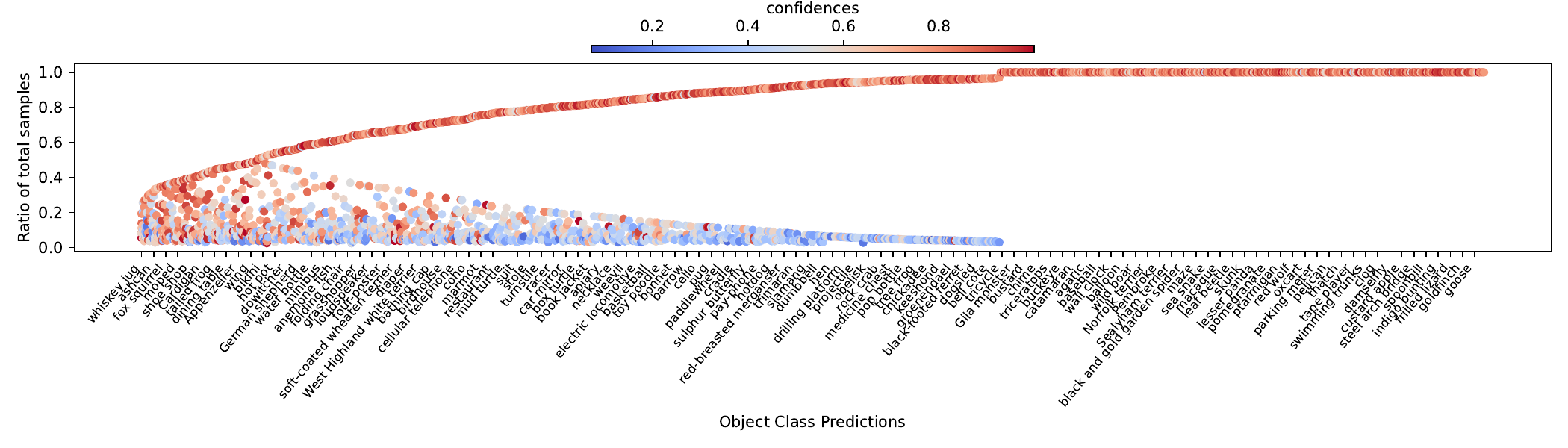}
  \caption{DenseNet121.}
  \label{fig:count_preds_conf_scatter_densenet121}
\end{figure}

\begin{figure}[ht]
  \centering
  \includegraphics[width=\linewidth]{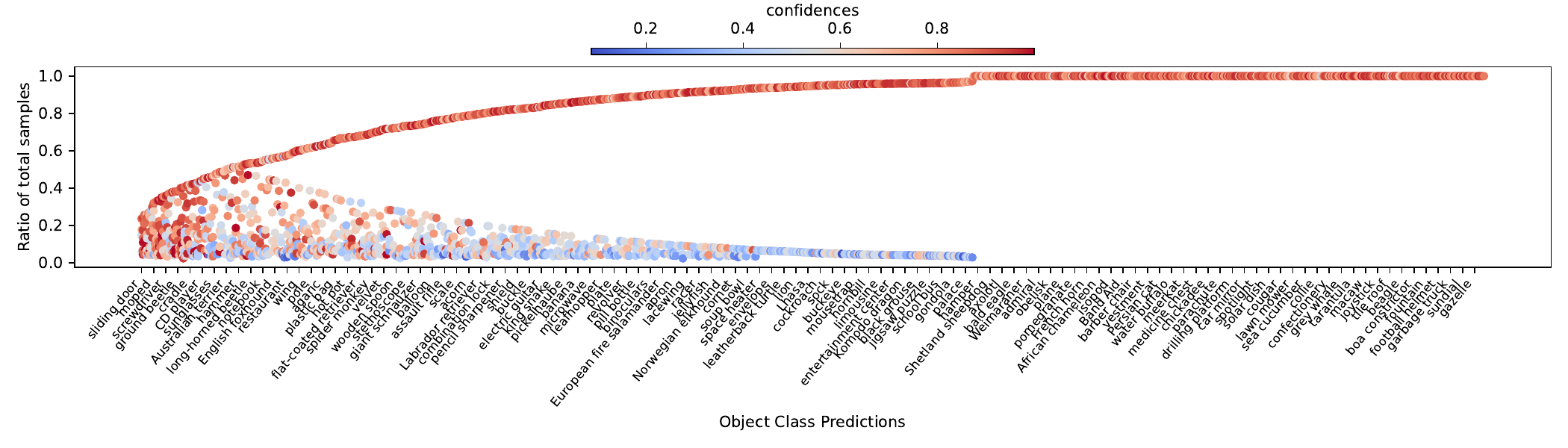}
  \caption{DenseNet169.}
  \label{fig:count_preds_conf_scatter_densenet169}
\end{figure}

\end{document}